\documentclass{article}

\usepackage{arxiv}
\usepackage{amssymb}
\usepackage{array}
\usepackage{arydshln}  %
\usepackage{amsfonts}       %
\usepackage{booktabs}       %
\usepackage[T1]{fontenc}    %
\usepackage{hyperref}       %
\usepackage[utf8]{inputenc} %
\usepackage{lipsum}
\usepackage{microtype}      %
\usepackage[frozencache]{minted}         %
\usepackage{multirow}       %
\usepackage{nicefrac}       %
\usepackage{subcaption}     %
\usepackage{todonotes}
\usepackage{url}            %
\usepackage{xspace}         %

\usepackage{xpatch}  %

\makeatletter
\AtBeginEnvironment{minted}{\dontdofcolorbox}
\def\dontdofcolorbox{\renewcommand\fcolorbox[4][]{##4}}
\xpatchcmd{\inputminted}{\minted@fvset}{\minted@fvset\dontdofcolorbox}{}{}
\xpatchcmd{\mintinline}{\minted@fvset}{\minted@fvset\dontdofcolorbox}{}{} %
\makeatother

\newcolumntype{L}[1]{>{\raggedright\let\newline\\\arraybackslash\hspace{0pt}}m{#1}}
\newcolumntype{C}[1]{>{\centering\let\newline\\\arraybackslash\hspace{0pt}}m{#1}}
\newcolumntype{R}[1]{>{\raggedleft\let\newline\\\arraybackslash\hspace{0pt}}m{#1}}

\usepackage[compress,numbers,sort]{natbib}
\newcommand{\pg}{\texttt{python\_graphs}\xspace}
\newcommand{\code}[1]{\texttt{#1}}

\newcommand{\dest}{\emph{dest}\xspace}
\newcommand{\src}{\emph{src}\xspace}

\makeatletter

\newcommand*{\tabminted@finalstrut}[1]{%
  \ifdim\prevdepth>0pt
    \ifdim\dp#1>\prevdepth
      \vskip\dimexpr(\dp#1)-\prevdepth\relax
    \fi
  \else
    \vskip\dimexpr(\dp#1)\relax
  \fi
}
\newcommand*{\@tabmintedend}{%
  \let\@finalstrut\tabminted@finalstrut
}
\makeatother

\title{A Library for Representing Python Programs as Graphs for Machine Learning}
\author{%
David Bieber\thanks{Correspondence to: \texttt{dbieber@google.com}} \\
Google Research\\
\And
Kensen Shi\\
Google Research\\
\And
Petros Maniatis\\
Google Research\\
\And
Charles Sutton\\
Google Research\\
\And
Vincent Hellendoorn\\
Carnegie Mellon University\\
\And
Daniel Johnson\\
Google Research\\
\And
Daniel Tarlow\\
Google Research\\
}

\begin{document}
\maketitle

\begin{abstract}
Graph representations of programs are commonly a central element of machine learning for code research.
We introduce an open source Python library \pg that applies static analysis to construct graph representations of Python programs suitable for training machine learning models.
Our library admits the construction of control-flow graphs, data-flow graphs, and composite ``program graphs'' that combine control-flow, data-flow, syntactic, and lexical information about a program.
We present the capabilities and limitations of the library, perform a case study applying the library to millions of competitive programming submissions, and showcase the library's utility for machine learning research.
\end{abstract}

\section{Introduction}

\setcounter{footnote}{0}
In this report we present \pg\footnote{\url{https://github.com/google-research/python-graphs}}, a Python library for constructing graph representations of Python programs for use in machine learning research.
This report details the capabilities and limitations of the library as they pertain to applying machine learning to source code.

A standard class of approaches in applying machine learning to code is to construct a graph representation of a program, and then to perform the analysis of interest on that graph representation, learning from a large dataset of labeled example programs.
Graph representations of programs used for machine learning include the abstract syntax tree (AST), control-flow graph (CFG), data-flow graphs, inter-procedural control-flow graph (ICFG), interval graph, and composite ``program graphs'' that encode information from multiple of the aforementioned graphs, possibly with additional program-derived data.

The \pg library directly allows for the construction of some of these graph types (e.g., control-flow graphs and composite program graphs) from arbitrary Python programs, and it provides tools that aid in constructing the others. It has been used successfully in a variety of machine learning for code publications, and we make it available as free and open source software to allow for broader use.

In Section~\ref{sec:background} we present an overview of the use of graph representations of code in machine learning. In Section~\ref{sec:capabilities-and-limitations} we describe the capabilities (Section~\ref{sec:capabilities}), possible extensions (Section~\ref{sec:possible-extensions}), and limitations (Section~\ref{sec:limitations}) of \pg.
Section~\ref{sec:use-cases} highlights the applications of \pg for machine learning research.
Section~\ref{sec:case-study} presents a case study applying \pg to 3.3 million programs from Project CodeNet \citep{codenet}.

\section{Background}
\label{sec:background}

\paragraph{Graph representations of code in machine learning}

Graph representations of source code are regularly used in machine learning research.
Most common among these is the abstract syntax tree.
Several works learn directly from ASTs \citep{deep-code-clone,code2vec,code2seq,tree-clones,novel-ast,structural-code,atom-ast,ast-transformer,modular-tree,infercode,deep-rl4fl,flat-asts,langnostic}
or produce an AST as output \citep{asn,tranx}, while
\citet{gfsa} learns to dynamically augment an AST with new edges useful for a downstream task.
Other works operate on a program's control-flow graph \citep{cnn-cfg-defects,func2vec,ipagnn,runtime-errors} or data-flow graph \citep{tpu-perf, graph-code-bert}, or joint control and data flow graph (CDFG) \citep{design2vec}.
A typical composite program graph uses an AST backbone with some subset of control-flow, data-flow, lexical, and syntactic information encoded as additional edges \citep{cpg,learning-to-represent-programs-with-graphs,devign,great-paper,ggnn-log-levels,hpg,buglab,wip-pg}.
\citet{mocktail} meanwhile uses AST, CFG, and program dependence graph (PDG) representations concurrently, without unifying them into a single graph.
\citet{codetrek} forms a graph via CodeQL queries to a database representing a program.
\citet{heat} forms a hypergraph, where edges can connect more than two nodes, containing again control-flow, data-flow, lexical, and syntactic information.
Still other representations include a program's interval graph \citep{interval-graph} or a graph formed from the pointers in the heap \citep{ggnn}.
A graph can also encode additional information, e.g., as in \citet{build-repair} which constructs a graph jointly representing code, a compiler error, and a build file.

Our work directly admits constructing control-flow graphs, performing data-flow analyses, and constructing certain composite program graphs from Python programs (Section~\ref{sec:capabilities}). It can also be extended for constructing interprocedural control-flow graphs, novel composite program graphs, additional data-flow graphs, or span-mapped graphs (Section~\ref{sec:possible-extensions}).

\paragraph{Tools for constructing graph representations}

We compare \pg with existing Python static analysis tools.
Tree-sitter \citep{tree-sitter} can build a concrete syntax tree for a given source file and update it incrementally as the source changes.
It supports over 40 languages including Python.
Our system must operate directly on the built-in Python AST rather than a language agnostic syntax tree.
CodeQL \citep{codeql} is a query language for source code. These queries admit searching for control-flow and data-flow paths in source code.
pycfg \citep{pycfg} generates control-flow graphs from Python source in a similar manner to \pg, but lacks support for certain language features like exceptions and generators. Scalpel \citep{scalpel} similarly generates control-flow graphs from Python and also performs additional static analyses, e.g., call graph construction. \pg performs data-flow analyses on top of its control-flow graphs, producing composite program graphs containing control-flow, data-flow, syntactic, and lexical information in one graph.

\section{Capabilities, Possible Extensions, and Limitations}
\label{sec:capabilities-and-limitations}

We provide a comprehensive overview of the capabilities of the \pg library,
a discussion of how \pg can enable still further capabilities (i.e. assisting in constructing the graph types not directly supported by the library today),
and a discussion of the library's limitations.

\begin{figure}
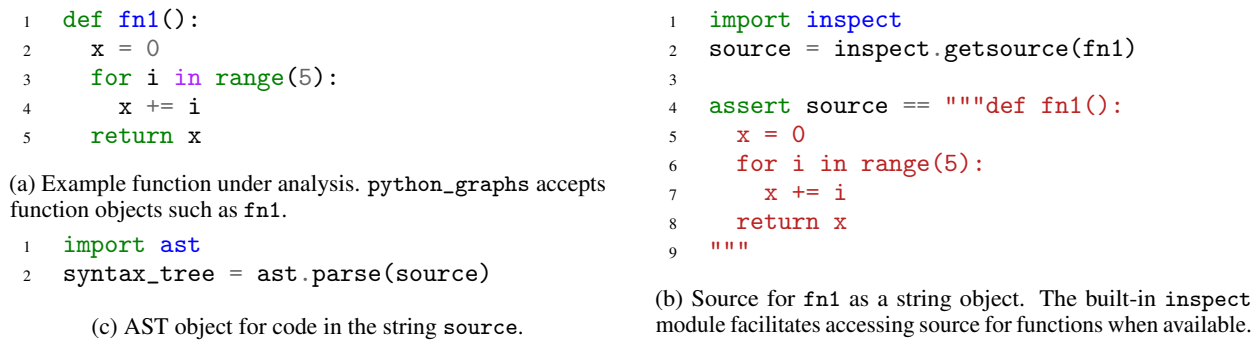

\begin{minipage}{0.48\textwidth}
\begin{subfigure}[t]{\textwidth}
\begin{minted}[xleftmargin=20pt,linenos]{python}
def fn1():
  x = 0
  for i in range(5):
    x += i
  return x
\end{minted}
\caption{Example function under analysis. \pg accepts function objects such as \texttt{fn1}.}
\end{subfigure}
\par\smallskip
\setcounter{subfigure}{2}
\begin{subfigure}[t]{\textwidth}
\begin{minted}[xleftmargin=20pt,linenos]{python}
import ast
syntax_tree = ast.parse(source)
\end{minted}
\caption{AST object for code in the string \texttt{source}.}
\end{subfigure}
\end{minipage}
\begin{minipage}{.03\textwidth}
\quad
\end{minipage}
\begin{minipage}{.48\textwidth}
\setcounter{subfigure}{1}
\begin{subfigure}[t]{\textwidth}
\begin{minted}[xleftmargin=20pt,linenos]{python}
import inspect
source = inspect.getsource(fn1)

assert source == """def fn1():
  x = 0
  for i in range(5):
    x += i
  return x
"""
\end{minted}
\caption{Source for \texttt{fn1} as a string object. The built-in \texttt{inspect} module facilitates accessing source for functions when available.}
\end{subfigure}
\end{minipage}
\caption{The input formats accepted by the \pg library are (a) function, (b) source code, and (c)  AST. The code snippets here demonstrate construction of each input format for the example function \texttt{fn1}.}
\label{fig:input-formats}
\end{figure}

\subsection{Capabilities}
\label{sec:capabilities}

The \pg library enables a number of static analyses on Python source code. The main use cases are computing control-flow graphs, performing data-flow analyses, constructing composite ``program graphs,'' and measuring cyclomatic complexity of Python programs and functions \citep{dragon-book}. Each of these operations may be applied to a full Python program or an individual Python function. The library handles any of the following input types: (a) Python function, (b) source code string, or (c) abstract syntax tree. Figure~\ref{fig:input-formats} shows constructing all three valid input formats for a sample program. In all cases, the library first converts the input to an abstract syntax tree for analysis.

\subsubsection{Control-Flow Graphs}

A control-flow graph represents the possible paths of execution through a program. Each node in a control-flow graph represents a basic block. A basic block is a straight-line section of source code that is executed contiguously. The only branches into a basic block enter at the start, and the only branches out of a basic block exit at the end (other than Exceptions). An edge in a control-flow graph represents a possible path of execution. There is an edge between node A and node B in a program's control-flow graph if and only if it is possible to execute basic block B immediately following the conclusion of executing basic block A \citep{dragon-book}.

\begin{figure}
\begin{minipage}{.57\textwidth}
\begin{subfigure}[t]{\textwidth}
\begin{minted}[xleftmargin=17pt,linenos]{python}
# 1. Use control_flow to construct a CFG.
from python_graphs import control_flow
graph = control_flow.get_control_flow_graph(fn1)
\end{minted}
\begin{minted}[xleftmargin=17pt,linenos]{python}
# 2. Access a particular basic block by source.
block = graph.get_block_by_source("x += i")
\end{minted}
\begin{minted}[xleftmargin=17pt,linenos]{python}
# 3. Convert the CFG to a pygraphviz.AGraph.
from python_graphs import control_flow_graphviz
agraph = control_flow_graphviz.to_graphviz(graph)
\end{minted}
\caption{Example usage of \pg to (1) construct a CFG, (2) access basic blocks by source, and (3) convert to pygraphviz for visualization.}
\label{fig:cfg-usage}
\end{subfigure}
\end{minipage}
\begin{minipage}{.03\textwidth}
\quad
\end{minipage}
\begin{minipage}{0.37\textwidth}
\begin{subfigure}[t]{\textwidth}
\begin{tabular}{rlr}
\toprule
n & Source & Control-Flow Graph \\
\midrule
1 &
\multirow{5}{6em}{%
\mintinline{python}{x = 0}\\
\mintinline{python}{.0 = range(5)}\\
\mintinline{python}{i = next(.0)}\\
\mintinline{python}{x += i}\\
\mintinline{python}{return x}\\
}
& 
\multirow{5}{*}{
\includegraphics[height=56pt]{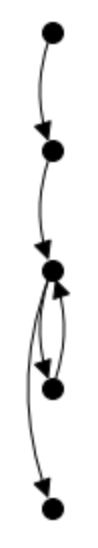}
}\\
 2 & & \\
 3 & & \\
 4 & & \\
 5 & & \\
\bottomrule
\end{tabular}
\caption{The statement-level control-flow graph for the program \texttt{fn1} introduced in Figure~\ref{fig:input-formats}. \texttt{.0} denotes the iterator constructed by the for loop in \texttt{fn1}.}
\label{fig:statement-level-cfg}
\end{subfigure}
\end{minipage}
\caption{\pg supports construction and analysis of control-flow graphs for arbitrary Python functions.}
\label{fig:cfg}
\end{figure}

In addition to producing these standard control-flow graphs, the \pg library can also produce statement-level control-flow graphs. A node in a statement-level control-flow graph represents a single line or instruction, rather than a complete basic block. An edge between two nodes indicates that the two lines may be executed in succession. Figure~\ref{fig:statement-level-cfg} shows the statement-level CFG for the \texttt{fn1} program.

A control-flow graph is useful for machine learning for source code in two respects.
First, it is a useful representation of code suitable for processing with graph neural networks, for example in 
\citet{ipagnn,runtime-errors}. Second, the control-flow graph forms the basis for a number of further analyses including data-flow analyses (liveness analysis, reachability, etc.), computing cyclomatic complexity, and constructing program graphs, each implemented by the \pg library.

In Python, any line of code can raise an exception. Taking this form of execution into account, this limits basic blocks to a single line of code, since a raised exception is an exit branch. Rather than restrict basic blocks to a single line of code, we take a more pragmatic approach, and introduce a second optional edge type, ``interrupting edges'', in our control-flow graph data structure that represents control flow due to exceptions. An interrupting edge from block A to block B indicates that an exception raised during the execution of A can cause control to flow to block B. \pg control-flow graphs can be used with or without these interrupting edges.

\setlength\dashlinedash{.5pt}
\setlength\dashlinegap{11pt}
\setlength\arrayrulewidth{0.3pt}

To construct a control-flow graph with \pg, use the \mintinline{python}{control_flow} module's \mintinline{python}{get_control_flow_graph} as in Figure~\ref{fig:cfg}.

\subsubsection{Data-Flow Analyses}

A data-flow analysis computes information about how the variables in the program are used, such as which variables are \emph{live} at a given program location. A live variable is one that may be read at some point in the future before its value is overwritten. The \pg library implements two best-effort data-flow analyses: liveness and last-access analysis.

Data-flow analyses are performed through iterative application of the data-flow equations until a fixed point is reached. The \pg library supports both forward and backward data-flow analysis, and so can be extended to support additional data-flow analyses. Liveness is implemented as a backward analysis, and last-access as a forward analysis.

An example of using the liveness analysis to obtain the set of loop variables in a while loop is provided with the library, a necessary step in rewriting Python while loops into their functional form. The \texttt{data\_flow} module provides the data-flow analyses, as in Figure~\ref{fig:data-flow}.

\subsubsection{Composite Program Graphs}
\label{sec:composite-graph}

The \pg library implements a single kind of composite program graph, based closely on that of \citet{learning-to-represent-programs-with-graphs}.
In this document we refer to these composite program graphs simply as ``program graphs'', though of course other kinds of program graphs are possible, with different node and edge types.

A program graph has the abstract syntax tree of the program it represents as its backbone. Each node in the program graph directly corresponds to a single node in the AST, and vice versa. Lists and primitive values in the AST have corresponding nodes in the program graph as well.
Corresponding to each syntax element in the program (leaf nodes in the AST) is a syntax node in the program graph.
Each edge in the AST also appears in the program graph. The program graph then has additional edges representing the following relationships between program pieces: ``NEXT\_SYNTAX'', ``LAST\_LEXICAL\_USE'', ``CFG\_NEXT'', ``LAST\_READ'', ``LAST\_WRITE'', ``COMPUTED\_FROM'', ``CALLS'', ``FORMAL\_ARG\_NAME'', and ``RETURNS\_TO''.
Collectively, the edges in a program graph convey control-flow, data-flow, lexical, and syntactic information about the program.

\begin{figure}
\begin{minipage}{.57\textwidth}
\begin{subfigure}[t]{\textwidth}
\begin{minted}[xleftmargin=17pt,linenos]{python}
# 1. Use data_flow to perform liveness analysis.
from python_graphs import data_flow
analysis = data_flow.LivenessAnalysis()
for block in graph.get_exit_blocks():
  analysis.visit(block)
\end{minted}
\begin{minted}[xleftmargin=17pt,linenos]{python}
# 2. Construct a program graph with program_graph.
from python_graphs import program_graph
pg = program_graph.get_program_graph(program)
\end{minted}
\begin{minted}[xleftmargin=17pt,linenos]{python}
# 3. Access a particular node by source.
node = pg.get_node_by_source_and_identifier(
    'return x', 'x')
\end{minted}
\caption{Example usage of \pg to (1) perform liveness analysis on a program's control-flow graph. Independently, (2) shows constructing a composite program graph, and (3) accessing one of its node by source.}
\label{fig:data-flow-usage}
\label{fig:program-graph-usage}
\end{subfigure}
\end{minipage}
\begin{minipage}{.03\textwidth}
\quad
\end{minipage}
\begin{minipage}{0.38\textwidth}
\begin{subfigure}[t]{\textwidth}
\begin{tabular}{clcc}
\toprule
\# & Source & Live in & Live out \\
\midrule
1
&
\multirow{5}{6em}{%
\mintinline{python}{x = 0}\\
\mintinline{python}{.0 = range(5)}\\
\mintinline{python}{i = next(.0)}\\
\mintinline{python}{x += i}\\
\mintinline{python}{return x}\\
}
&  $\varnothing$ &  \\
   & &  & \{\texttt{x}\} \\
 2 & & \{\texttt{x}\} & \{\texttt{x}, \texttt{i}\} \\
 3 & & \{\texttt{x}, \texttt{i}\} & \{\texttt{x}\} \\
 4 & & \{\texttt{x}\} & $\varnothing$ \\
\bottomrule
\end{tabular}
\caption{The results of the liveness data-flow analysis. Live in and live out indicate the variables that are live at the start and end of each basic block respectively. \# denotes basic block number.}
\label{fig:data-flow-results}
\end{subfigure}
\end{minipage}
\caption{Example usage of data-flow analysis and program graph construction in \pg.}
\label{fig:data-flow}
\end{figure}

We summarize the edge types and their meanings in Table~\ref{tab:edge-descriptions}. These edge types are also useful for constructing other graph types (Section~\ref{sec:possible-extensions}): interprocedural control-flow graphs, data-flow graphs, and other composite program graphs.

\begin{table}[b!]
    \centering
    \small
    \begin{tabular}{rp{12.2cm}}
\toprule
\sc{Edge type}  & \sc{Description} \\
\midrule
\sc{FIELD}                &  \dest is a field of AST node \src. \\
\sc{NEXT\_SYNTAX}         &  \dest is the syntax element immediately following \src in top-to-bottom left-to-right order. \\
\sc{LAST\_LEXICAL\_USE}   &  The variable at node \dest has its previous appearance at \src in top-to-bottom left-to-right order. \\
\sc{CFG\_NEXT}            &  The statement indicated by \dest can be executed immediately following that indicated by \src. \\
\sc{LAST\_READ}           &  When \src is about to execute, it may be that the variable at \src was most recently read at \dest. \\
\sc{LAST\_WRITE}           &  When \src is about to execute, it may be that the variable at \src was most recently written to at \dest. \\
\sc{COMPUTED\_FROM}       &  \src indicates a variable on an assignment's left hand side, and \dest a variable on its right hand side. \\
\sc{CALLS}                &  \src is an AST call node, and \dest is the definition of the function being called. \\
\sc{FORMAL\_ARG\_NAME}    &  \src is an argument in a function call; \dest is the corresponding parameter in the function definition. \\
\sc{RETURNS\_TO}          &  \src is the return node in a function definition, and \dest is the AST call node that calls that function. \\
\bottomrule
    \end{tabular}
    \caption{An edge of the given edge type from node \src to node \dest has the meaning given in this table.}
    \label{tab:edge-descriptions}
\end{table}

The \pg library provides the function \code{get\_program\_graph} for constructing a program graph from any of the supported input types (source code, an abstract syntax tree, or Python function). Figure~\ref{fig:program-graph-usage} shows example usage.

Table~\ref{tab:table-of-graphs} lists several programs along with their control-flow graphs and program graphs as computed by \pg.

\begin{table}[hp!]
\centering
\begin{tabular}{ll|rp{4.2cm}cC{3cm}}
\toprule
\# & Source & n & Statement & CFG & Program Graph\\
\midrule
  \multirow{3}{0.75em}{1} &
  \multirow{3}{15em}{%
    \mintinline{python}{def fn1(a, b):}\\
  ~~~~\mintinline{python}{return a + b}\\
  \mintinline{python}{}\\
  }&
  1 &
  \multirow{3}{9em}{%
    \mintinline{python}{a, b ← args}\\
  \mintinline{python}{return a + b}\\
  \mintinline{python}{<exit>}\\
  } &
  \multirow{3}{1em}{%
  \centering
  \includegraphics[height=33pt]{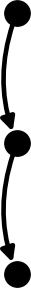}
  } &
  \multirow{3}{7em}{%
  \centering
  \includegraphics[height=22pt]{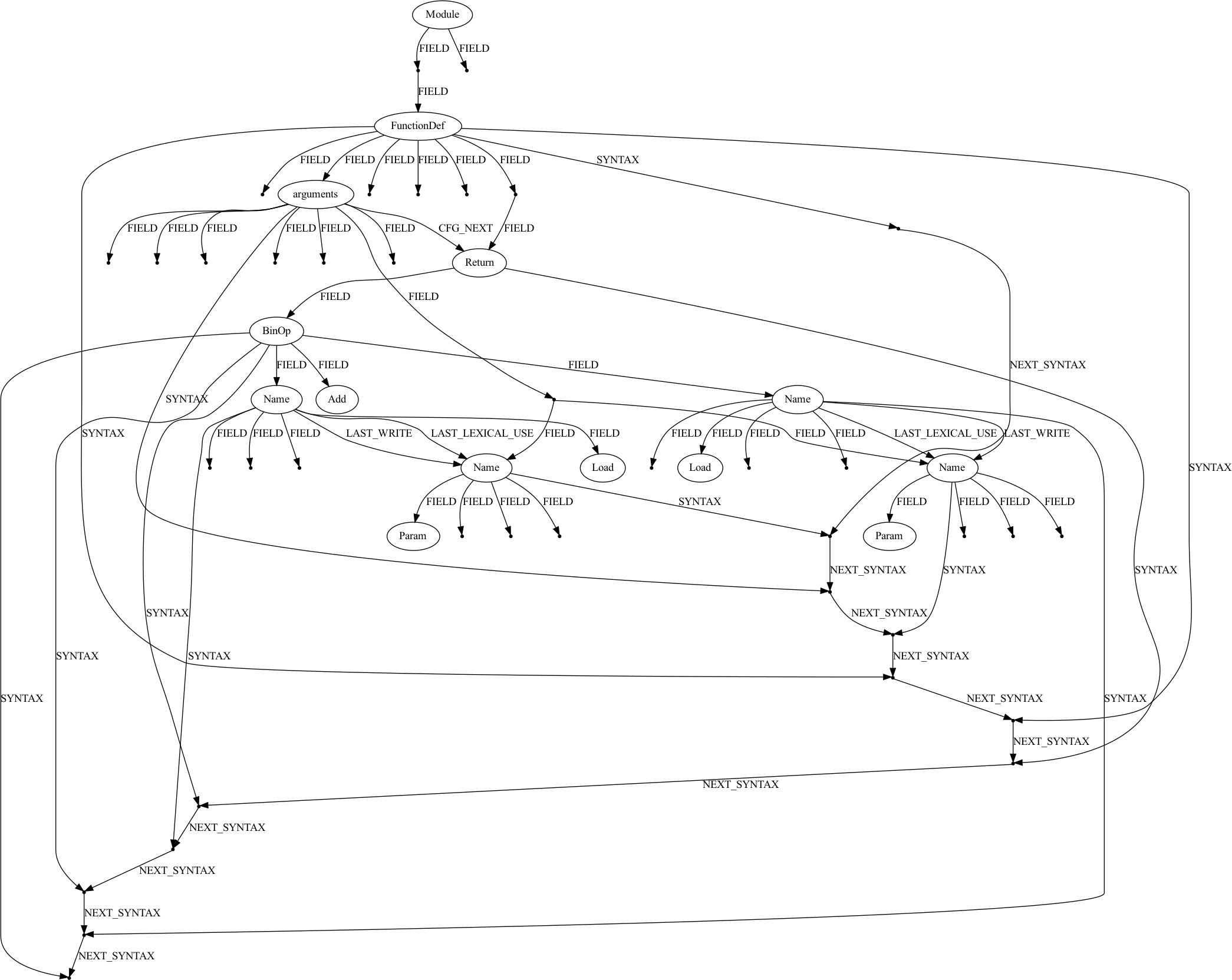}\\
  \begin{small}Figure~\ref{fig:program-graph-1}\end{small}
  }
  \\
   & & 2 & & \\
 & & 3 & & \\
\midrule
  \multirow{6}{0.75em}{2} &
  \multirow{6}{15em}{%
    \mintinline{python}{def fn2(a, b):}\\
  ~~~~\mintinline{python}{c = a}\\
  ~~~~\mintinline{python}{if a > b:}\\
  ~~~~~~~~\mintinline{python}{c -= b}\\
  ~~~~\mintinline{python}{return c}\\
  \mintinline{python}{}\\
  }&
  1 &
  \multirow{6}{9em}{%
    \mintinline{python}{a, b ← args}\\
  \mintinline{python}{c = a}\\
  \mintinline{python}{a > b}\\
  \mintinline{python}{c -= b}\\
  \mintinline{python}{return c}\\
  \mintinline{python}{<exit>}\\
  } &
  \multirow{6}{1em}{%
  \centering
  \includegraphics[height=65pt]{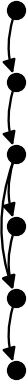}
  } &
  \multirow{6}{7em}{%
  \centering
  \includegraphics[height=56pt]{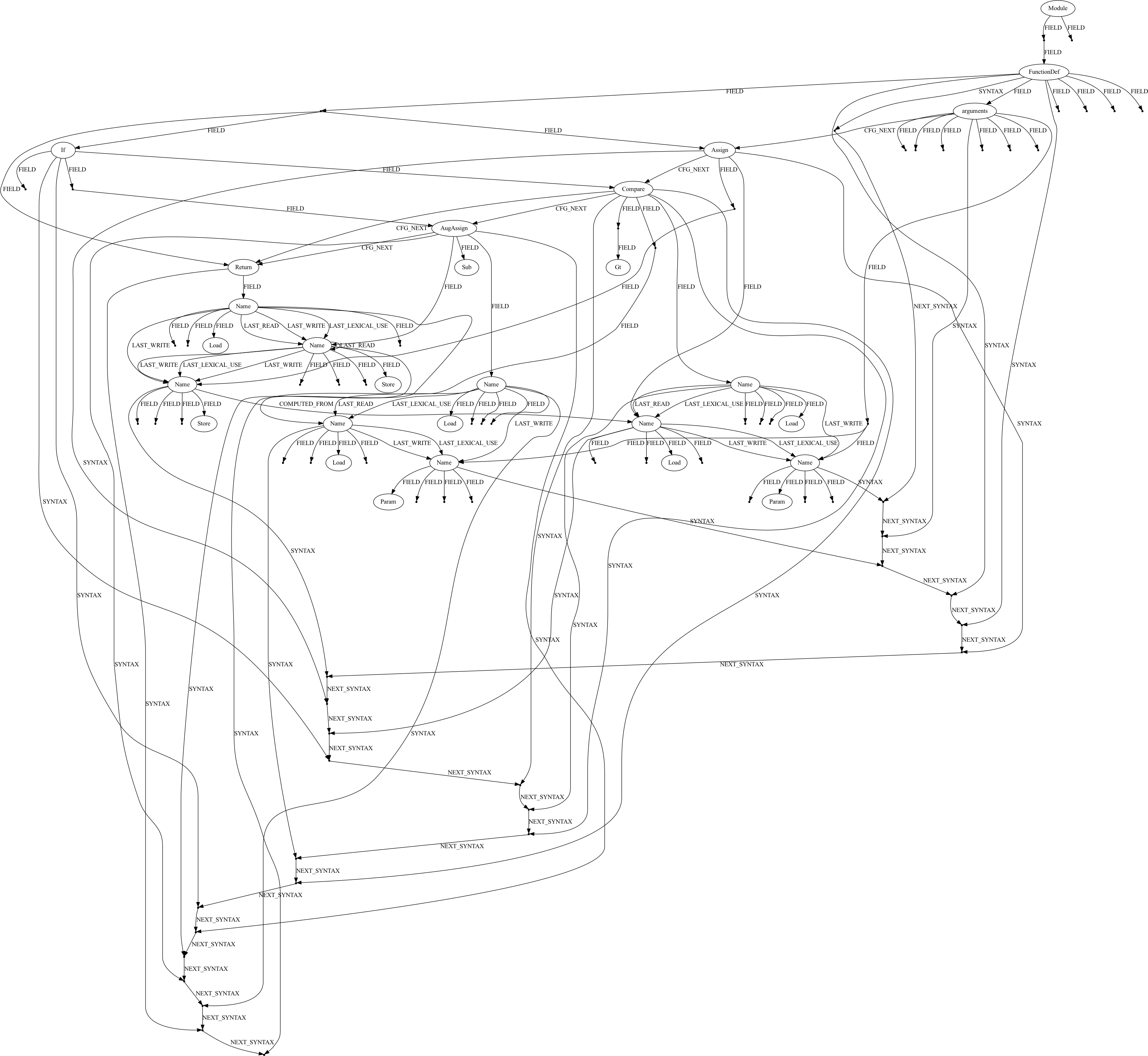}\\
  \begin{small}Figure~\ref{fig:program-graph-2}\end{small}
  }
  \\
   & & 2 & & \\
 & & 3 & & \\
 & & 4 & & \\
 & & 5 & & \\
 & & 6 & & \\
\midrule
  \multirow{10}{0.75em}{3} &
  \multirow{10}{15em}{%
    \mintinline{python}{def fn3(a, b):}\\
  ~~~~\mintinline{python}{c = a}\\
  ~~~~\mintinline{python}{if a > b:}\\
  ~~~~~~~~\mintinline{python}{c -= b}\\
  ~~~~~~~~\mintinline{python}{c += 1}\\
  ~~~~~~~~\mintinline{python}{c += 2}\\
  ~~~~~~~~\mintinline{python}{c += 3}\\
  ~~~~\mintinline{python}{else:}\\
  ~~~~~~~~\mintinline{python}{c += b}\\
  ~~~~\mintinline{python}{return c}\\
  }&
  1 &
  \multirow{10}{9em}{%
    \mintinline{python}{a, b ← args}\\
  \mintinline{python}{c = a}\\
  \mintinline{python}{a > b}\\
  \mintinline{python}{c -= b}\\
  \mintinline{python}{c += 1}\\
  \mintinline{python}{c += 2}\\
  \mintinline{python}{c += 3}\\
  \mintinline{python}{c += b}\\
  \mintinline{python}{return c}\\
  \mintinline{python}{<exit>}\\
  } &
  \multirow{10}{1em}{%
  \centering
  \includegraphics[height=110pt]{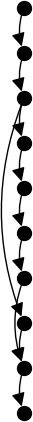}
  } &
  \multirow{10}{7em}{%
  \centering
  \includegraphics[height=100pt]{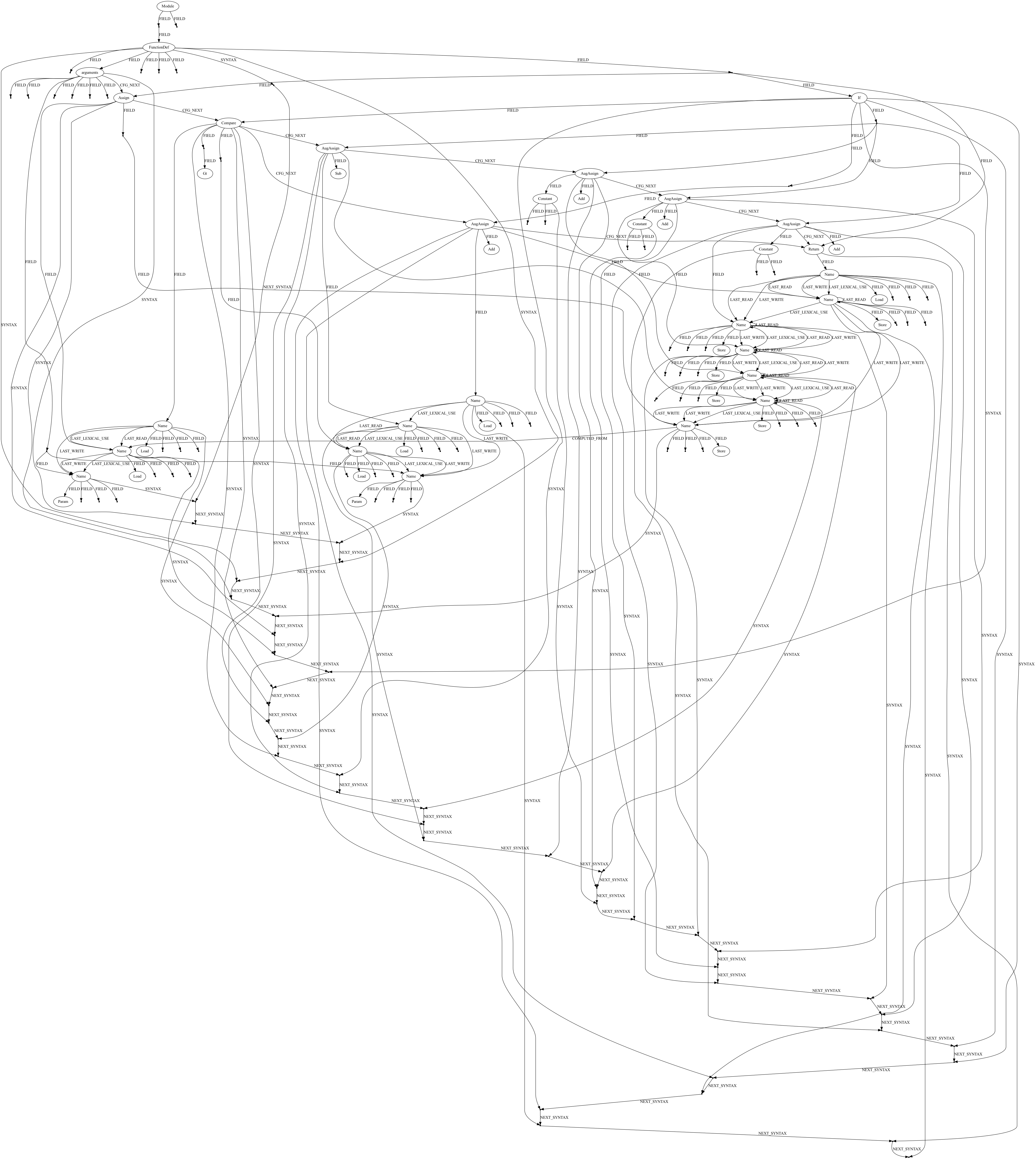}\\
  \begin{small}Figure~\ref{fig:program-graph-3}\end{small}
  }
  \\
   & & 2 & & \\
 & & 3 & & \\
 & & 4 & & \\
 & & 5 & & \\
 & & 6 & & \\
 & & 7 & & \\
 & & 8 & & \\
 & & 9 & & \\
 & & 10 & & \\
\midrule
  \multirow{7}{0.75em}{4} &
  \multirow{7}{15em}{%
    \mintinline{python}{def fn4(i):}\\
  ~~~~\mintinline{python}{count = 0}\\
  ~~~~\mintinline{python}{for i in range(i):}\\
  ~~~~~~~~\mintinline{python}{count += 1}\\
  ~~~~\mintinline{python}{return count}\\
  \mintinline{python}{}\\
  \mintinline{python}{}\\
  }&
  1 &
  \multirow{7}{9em}{%
    \mintinline{python}{i ← args}\\
  \mintinline{python}{count = 0}\\
  \mintinline{python}{range(i)}\\
  \mintinline{python}{i ← iter}\\
  \mintinline{python}{count += 1}\\
  \mintinline{python}{return count}\\
  \mintinline{python}{<exit>}\\
  } &
  \multirow{7}{1em}{%
  \centering
  \includegraphics[height=76pt]{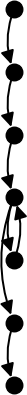}
  } &
  \multirow{7}{7em}{%
  \centering
  \includegraphics[height=67pt]{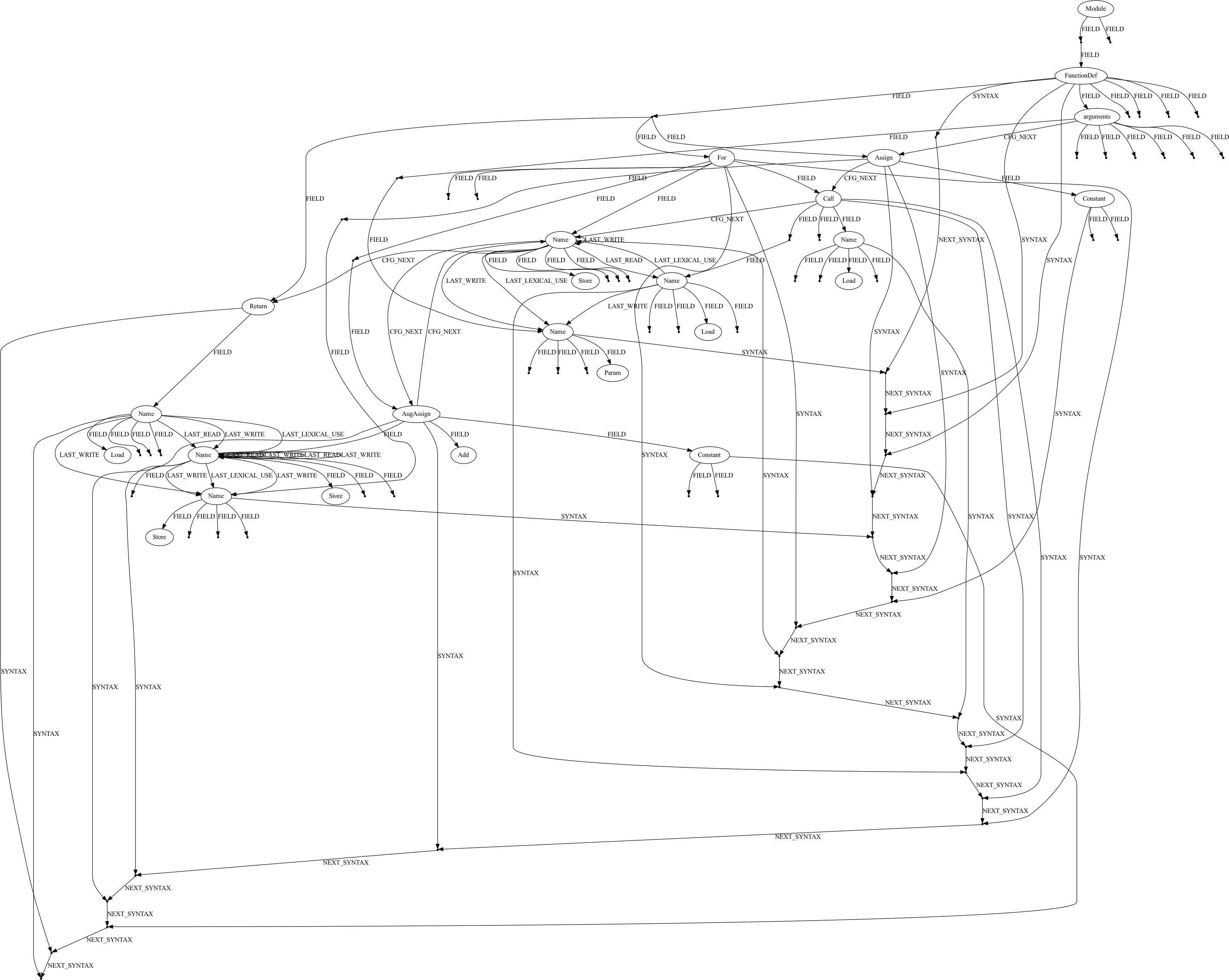}\\
  \begin{small}Figure~\ref{fig:program-graph-4}\end{small}
  }
  \\
   & & 2 & & \\
 & & 3 & & \\
 & & 4 & & \\
 & & 5 & & \\
 & & 6 & & \\
 & & 7 & & \\
\midrule
  \multirow{8}{0.75em}{5} &
  \multirow{8}{15em}{%
    \mintinline{python}{def fn5(i):}\\
  ~~~~\mintinline{python}{count = 0}\\
  ~~~~\mintinline{python}{for _ in range(i):}\\
  ~~~~~~~~\mintinline{python}{if count > 5:}\\
  ~~~~~~~~~~~~\mintinline{python}{break}\\
  ~~~~~~~~\mintinline{python}{count += 1}\\
  ~~~~\mintinline{python}{return count}\\
  \mintinline{python}{}\\
  }&
  1 &
  \multirow{8}{9em}{%
    \mintinline{python}{i ← args}\\
  \mintinline{python}{count = 0}\\
  \mintinline{python}{range(i)}\\
  \mintinline{python}{_ ← iter}\\
  \mintinline{python}{count > 5}\\
  \mintinline{python}{count += 1}\\
  \mintinline{python}{return count}\\
  \mintinline{python}{<exit>}\\
  } &
  \multirow{8}{1em}{%
  \centering
  \includegraphics[height=87pt]{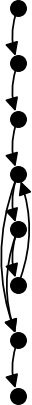}
  } &
  \multirow{8}{7em}{%
  \centering
  \includegraphics[height=78pt]{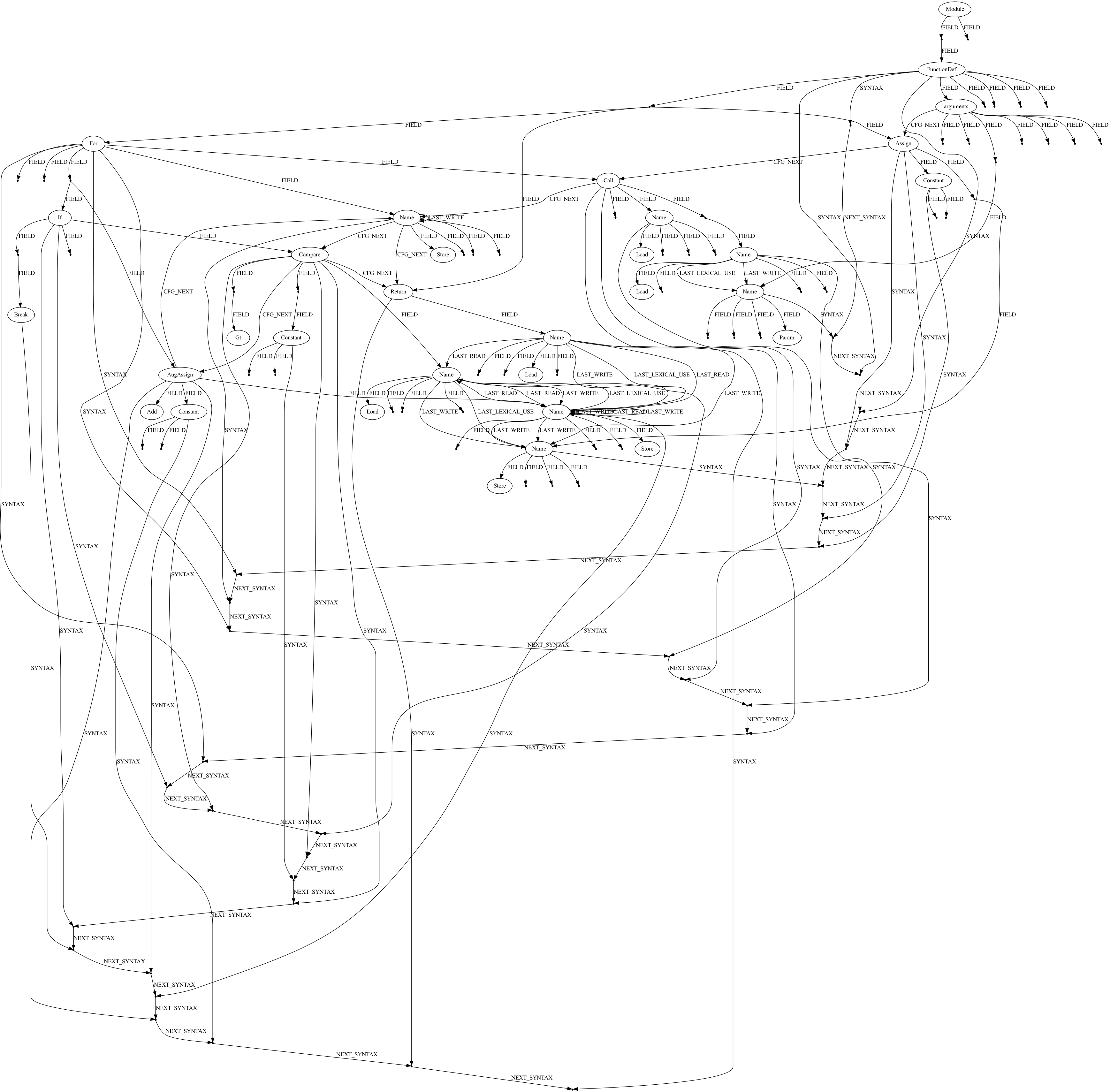}\\
  \begin{small}Figure~\ref{fig:program-graph-5}\end{small}
  }
  \\
   & & 2 & & \\
 & & 3 & & \\
 & & 4 & & \\
 & & 5 & & \\
 & & 6 & & \\
 & & 7 & & \\
 & & 8 & & \\
\midrule
  \multirow{5}{0.75em}{6} &
  \multirow{5}{15em}{%
    \mintinline{python}{def fn6():}\\
  ~~~~\mintinline{python}{count = 0}\\
  ~~~~\mintinline{python}{while count < 10:}\\
  ~~~~~~~~\mintinline{python}{count += 1}\\
  ~~~~\mintinline{python}{return count}\\
  }&
  1 &
  \multirow{5}{9em}{%
    \mintinline{python}{count = 0}\\
  \mintinline{python}{count < 10}\\
  \mintinline{python}{count += 1}\\
  \mintinline{python}{return count}\\
  \mintinline{python}{<exit>}\\
  } &
  \multirow{5}{1em}{%
  \centering
  \includegraphics[height=54pt]{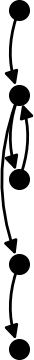}
  } &
  \multirow{5}{7em}{%
  \centering
  \includegraphics[height=44pt]{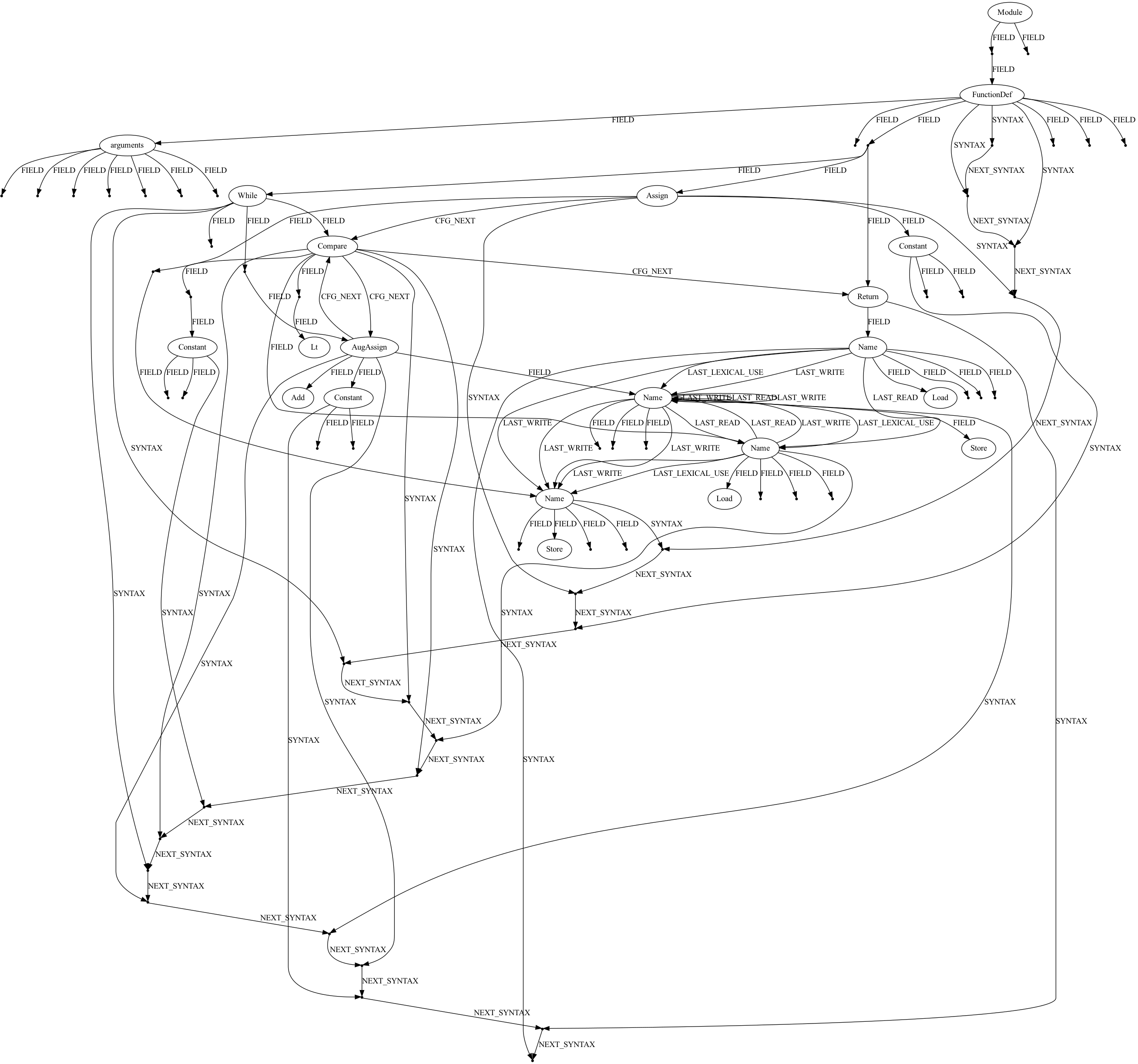}\\
  \begin{small}Figure~\ref{fig:program-graph-6}\end{small}
  }
  \\
   & & 2 & & \\
 & & 3 & & \\
 & & 4 & & \\
 & & 5 & & \\
\midrule
  \multirow{6}{0.75em}{7} &
  \multirow{6}{15em}{%
    \mintinline{python}{def fn7():}\\
  ~~~~\mintinline{python}{try:}\\
  ~~~~~~~~\mintinline{python}{raise ValueError('N/A')}\\
  ~~~~\mintinline{python}{except ValueError as e:}\\
  ~~~~~~~~\mintinline{python}{del e}\\
  ~~~~\mintinline{python}{return}\\
  }&
  1 &
  \multirow{6}{9em}{%
    \mintinline{python}{raise ValueError('N/A')}\\
  \mintinline{python}{ValueError}\\
  \mintinline{python}{e ← exception}\\
  \mintinline{python}{del e}\\
  \mintinline{python}{return}\\
  \mintinline{python}{<exit>}\\
  } &
  \multirow{6}{1em}{%
  \centering
  \includegraphics[height=65pt]{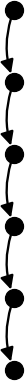}
  } &
  \multirow{6}{7em}{%
  \centering
  \includegraphics[height=56pt]{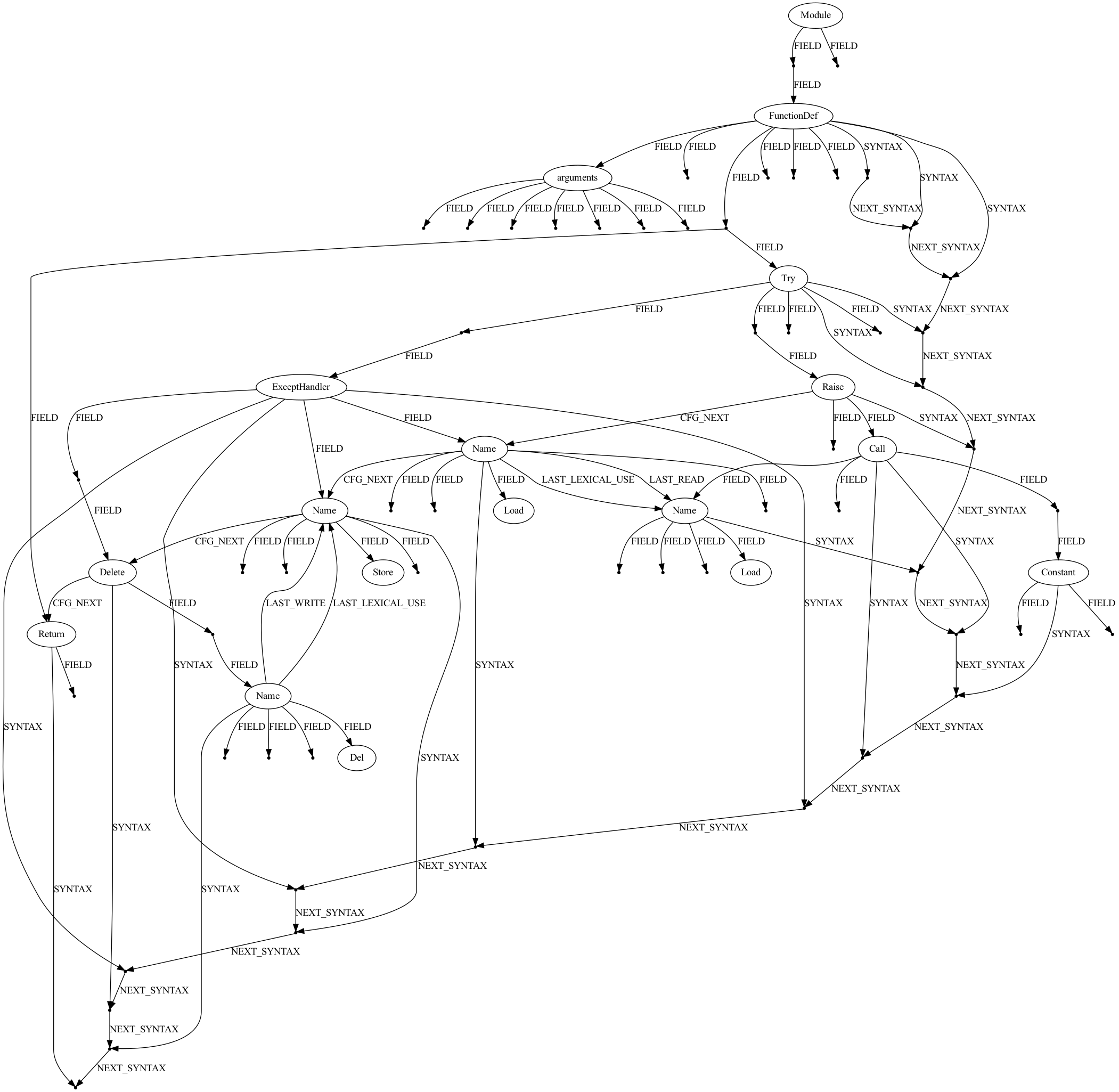}\\
  \begin{small}Figure~\ref{fig:program-graph-7}\end{small}
  }
  \\
   & & 2 & & \\
 & & 3 & & \\
 & & 4 & & \\
 & & 5 & & \\
 & & 6 & & \\
\midrule
  \multirow{3}{0.75em}{8} &
  \multirow{3}{15em}{%
    \mintinline{python}{def fn8(a):}\\
  ~~~~\mintinline{python}{a += 1}\\
  \mintinline{python}{}\\
  }&
  1 &
  \multirow{3}{9em}{%
    \mintinline{python}{a ← args}\\
  \mintinline{python}{a += 1}\\
  \mintinline{python}{<exit>}\\
  } &
  \multirow{3}{1em}{%
  \centering
  \includegraphics[height=33pt]{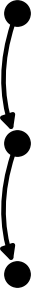}
  } &
  \multirow{3}{7em}{%
  \centering
  \includegraphics[height=22pt]{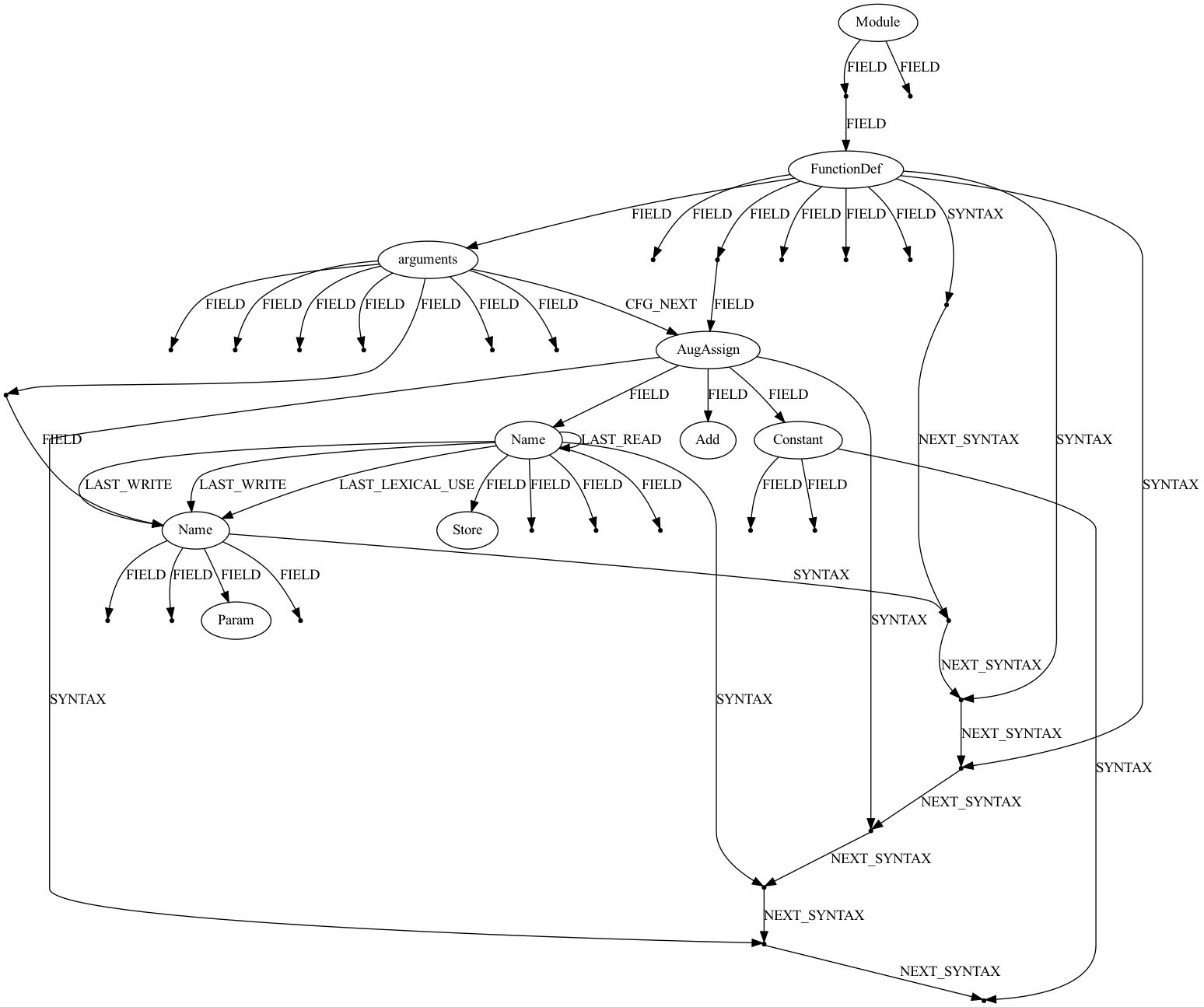}\\
  \begin{small}Figure~\ref{fig:program-graph-8}\end{small}
  }
  \\
   & & 2 & & \\
 & & 3 & & \\
\bottomrule
\end{tabular}
    \caption{Example programs and their associated control-flow graphs and program graphs. Enlarged versions of the program graph figures are included in Appendix~\ref{app:program-graph-examples}.
    }
    \label{tab:table-of-graphs}
\end{table}

\subsubsection{Cyclomatic Complexity}

Cyclomatic complexity is a standard measure of program complexity based on the set of possible paths through a program.
It measures the number of linearly independent execution paths through a program.
The \pg library can compute the cyclomatic complexity of a Python function.
This functionality is available via the function \code{cyclomatic\_complexity}, which accepts a Python function (as source, AST, or Python function object) and returns an integer.
To compute the cyclomatic complexity of a program, \pg first constructs its control-flow graph.
In a control-flow graph with $E$ edges, $N$ nodes, and $P$ distinct connected components, the cyclomatic complexity $M$ is given by $M = E - N + 2P$.

\subsection{Possible Extensions}
\label{sec:possible-extensions}

The following capabilities are possible to implement using \pg, but are not directly provided by \pg out of the box.

\subsubsection{Alternative Composite Program Graphs}
\label{sec:augmented-graph}

The program graphs generated by \pg's \code{get\_program\_graph} make specific choices for what nodes and edges are included in the graph. Other choices are possible. For alternative composite program graph schemes, the source of \pg's \code{get\_program\_graph} function serves as an illustrative example for how to construct a composite program graph with the desired set of nodes and edges.

\subsubsection{Inter-procedural Control-Flow Graphs}

\pg's \code{get\_control\_flow\_graph} function constructs the control-flow graph for a single function or program; it does not include edges indicating control flow between functions.

An interprocedural control-flow graph (ICFG) is a control-flow graph that shows the control flow possible between functions, not just within a function. We can view an ICFG as a composite program graph consisting of the control-flow graphs of a program and all its functions, as well as CALLS and RETURNS\_TO edges indicating the possible interprocedural control flows in the program. \pg provides the capability for constructing the necessary control-flow graphs and additional edges, making it possible to write a function to construct ICFGs as well.

\subsubsection{Data-Flow Graphs}

A data-flow graph represents the data dependencies present in a program.
The nodes in a data-flow graph represent the variable access locations in a program, and the edges in a data-flow graph denote relationships between these accesses.
An example of such a relationship is an edge with \dest indicating where a variable is assigned to and \src indicating where the assigned value is subsequently used (equivalent to \pg's LAST\_WRITE edges).
We can therefore view data-flow graphs as composite program graphs consisting of a subset of AST nodes (just those representing variable accesses) and selected edge types like LAST\_READ or LAST\_WRITE.

\subsubsection{Span-Mapped Graphs}
\label{sec:span-map}

In order to use the graphs produced by the \pg library for machine learning applications, it can be useful to tokenize the sections of code corresponding to each node. We suggest two approaches to handling this: (1) whole program tokenization and (2) per-node tokenization.

In \emph{whole program tokenization} we tokenize the entire program first. 
Then, using \pg, we create a graph structure for the program. 
Finally we extract for each node the span of tokens from the whole program tokenization corresponding to that node.
This approach allows for the possibility that a token consisting of multiple characters will be part of two consecutive nodes, and we must choose which node(s) to associate that token with.

In \emph{per-node tokenization}, we instead split the program source into chunks according to which node in the graph they are part of, and then tokenize those chunks independently.

The key data required by these approaches is a mapping from a graph node to a span in the textual representation of program source (approaches 1 and 2) or to a span in the tokenized representation of program source (approach 1 only). We call a graph augmented with this data a \emph{span-mapped graph}.
Both approaches are possible using \pg. \citet{runtime-errors} uses approach 1, with code freely available online\footnote{https://github.com/google-research/runtime-error-prediction}. This same code example is informative for any project wishing to implement approach 2.

\subsubsection{Additional Data-Flow Analyses}

\pg implements liveness and last-access data-flow analyses, and provides a  framework for implementing additional analyses.
This framework allows somewhat straightforward implementation of
definite assignment analysis
or computing reaching definitions, for example.

\subsection{Limitations}
\label{sec:limitations}

Python source code is difficult to analyze statically because so much of Python's behavior is determined dynamically.
We perform a ``best-effort'' analysis, which we do not guarantee will be correct considering all of Python's language features.
Inspection in Python allows manipulation of stack frames or of local or global variables, causing hard-to-detect effects on data and control flow.
Eval and exec permit the execution of arbitrary code constructed dynamically and inaccessible to our analysis.
Operations can be overloaded dynamically, so e.g. even a simple addition operation can have effects overlooked by our analyses.
These language features are empowering to Python users, but restrict the guarantees our analyses can provide.

\section{Use cases}
\label{sec:use-cases}

We next show how the \pg library is used in machine learning research.
Uses include both building graph representations of programs as inputs to neural networks, and providing supervision for models that output graphs.

\subsection{Graph Representations as Model Inputs}

\paragraph{Instruction Pointer Attention Graph Neural Networks}
The instruction pointer attention graph neural network (IPA-GNN) model family \citep{ipagnn, runtime-errors} operates on control-flow graphs as its primary input. IPA-GNN architectures then perform a soft execution of the input program, maintaining a soft instruction pointer representing at each model step a probability distribution over the statements in a program. These works use \pg to produce the control-flow graphs for the programs under consideration, which include both simple synthetic programs \citep{ipagnn} and complex human-authored programs from a competition setting \citep{runtime-errors}.

The original IPA-GNN work \citep{ipagnn} uses the control-flow edges as produced by \pg's default settings,
and represents each statement with a 4-tuple of values, which is possible because the domain of statements is restricted.
By contrast, the follow-up work on competition programs \citep{runtime-errors} uses a larger control-flow graph that additionally includes interrupting edges, indicating to where control would flow from each node if an exception were raised during the execution of that node.
Further, a sequence of tokens is associated with each node in the control-flow graph, following Section~\ref{sec:span-map}, allowing it to handle arbitrary human-authored Python statements.

\paragraph{Global Relational Models of Source Code}
\citet{great-paper} investigates models that combine global information (like a Transformer) and structural information (like a GNN), i.e.,
Graph-Sandwich models and the GREAT (Graph-Relational Embedding Attention Transformer) model. This paper uses \pg to construct the composite program graphs of Section~\ref{sec:composite-graph}. The models accept these program graphs as input and uses them to identify variable misuse bugs.

\subsection{Program Graphs as Targets}
\paragraph{Graph Finite-State Automaton (GFSA) Layers}
\citet{gfsa} introduces a neural network layer that adds a new learned edge type to an input graph.
Toward learning static analyses of Python code, it trains a neural model to take an AST as input and predict a composite program graph as output.
The model thereby learns to perform both control-flow and data-flow analyses from data.
For its targets, it produces a composite program graph with \pg, selecting a subset of the default edge types and introducing a few additional edge types (as in Section~\ref{sec:augmented-graph}).

\paragraph{Learning to Extend Program Graphs to Work-in-Progress Code}

\citet{wip-pg} learns to predict edge relations from work-in-progress code, even when the code does not parse. The composite program graphs of Section~\ref{sec:composite-graph} form the ground truth edge relation targets for this work.

\section{Case Study: Project CodeNet}
\label{sec:case-study}

In order to evaluate the \pg library on the diversity of language features found in realistic code, we obtain a dataset of 3.3 million programs from Project CodeNet \citep{codenet}. For each program, we use \pg to construct a control-flow graph and a composite program graph complete with syntactic, control-flow, data-flow, and lexical information about the program. We collect metrics about the resulting graphs to provide information about the robustness of \pg and the size, complexity, and connectedness of the program graphs it produces.

\begin{table}[t!]
\begin{minipage}[h]{0.49\textwidth}
    \centering
    \begin{tabular}{lrr}
    \toprule
    \textbf{Status}  & \textbf{\# Programs} & \textbf{Freq. (\%)}  \\
    \midrule
    Success                                      & 3,157,463    & 96.08     \\
    \mintinline{python}{ast.parse} failures      & 126,751     & 3.86      \\
    \quad SyntaxError                            & 114,817     & 3.49      \\
    \quad IndentationError                       & 8,893       & 0.27      \\
    \quad TabError                               & 3,032       & 0.09     \\
    \quad RecursionError                         & 5          & 0.00      \\
    \quad ValueError                             & 4          & 0.00      \\
    RuntimeError                                 & 2,100       & 0.06     \\
    \quad return outside function                & 1,719       & 0.05     \\
    \quad break outside loop                     & 330        & 0.01     \\
    \quad continue outside loop                  & 51         & 0.00    \\
    \midrule
    \textbf{Total}             & 3,286,314   & 100\%              \\
    \bottomrule
    \end{tabular}
    \caption{Control-flow graph construction success rates on a dataset of both valid and invalid Python submissions to competitive programming problems.}
    \label{tab:graph-status}
    \vspace{6pt}  %
\end{minipage}%
\hspace{0.019\textwidth}%
\begin{minipage}[h]{0.49\textwidth}
    \centering
    \begin{tabular}{rrr}
        \toprule
        \textbf{Edge Type} & \textbf{\# / Program} & \textbf{Freq. (\%)}   \\
        \midrule
        \small FIELD                       & 370.4   & 100.00 \\
        \small SYNTAX                      & 163.4   & 99.99 \\
        \small NEXT\_SYNTAX                & 162.4   & 99.99 \\
        \small LAST\_LEXICAL\_USE          & 26.1    & 99.12 \\
        \small CFG\_NEXT                   & 18.1    & 99.06 \\
        \small LAST\_READ                  & 38.0    & 92.29 \\
        \small LAST\_WRITE                 & 30.6    & 98.83 \\
        \small COMPUTED\_FROM              & 11.7    & 98.55 \\
        \small CALLS                       & 0.5     & 21.37 \\
        \small FORMAL\_ARG\_NAME           & 0.6     & 12.99 \\
        \small RETURNS\_TO                 & 0.7     & 15.56 \\
     \bottomrule
    \end{tabular}
    \caption{Frequencies of edge types in the composite program graphs for the Project CodeNet Python submissions. \# / Program is the average number of occurrences of the edge type across all programs. Freq. (\%) shows the percent of programs exhibiting the edge type at least once.}
    \label{tab:type-frequencies}
\end{minipage}
\end{table}

\begin{figure}[b!]
\centering
\begin{minipage}{.49\textwidth}
    \centering
    \includegraphics[width=\textwidth,trim=30pt 10pt 70pt 55pt,clip]{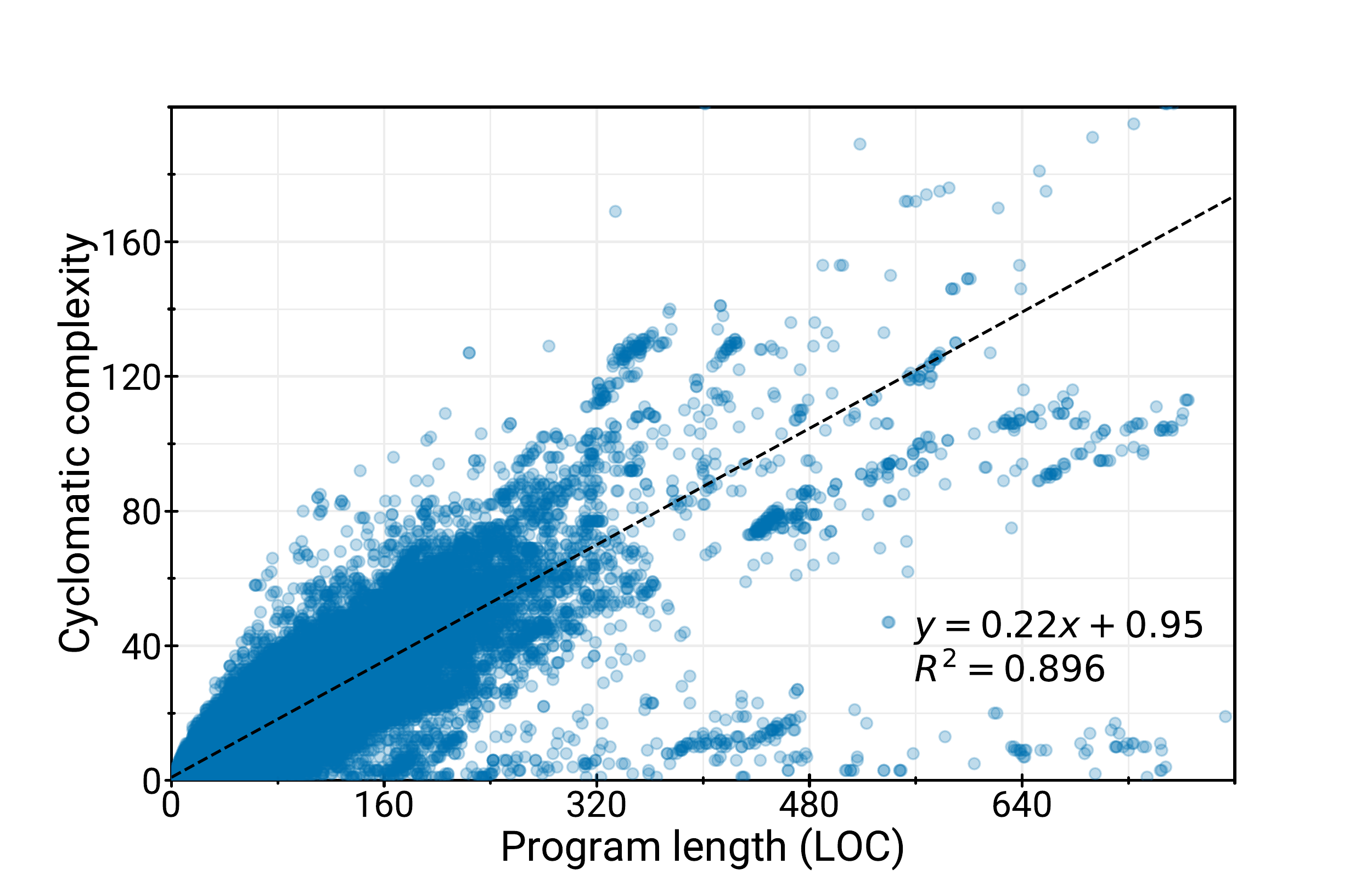}
    \caption{The relationship between program length and cyclomatic complexity for Python submissions in Project CodeNet.}
    \label{fig:cyclomatic-length}
\end{minipage}%
\hspace{0.019\textwidth}%
\begin{minipage}{.49\textwidth}
    \centering
    \includegraphics[width=\textwidth]{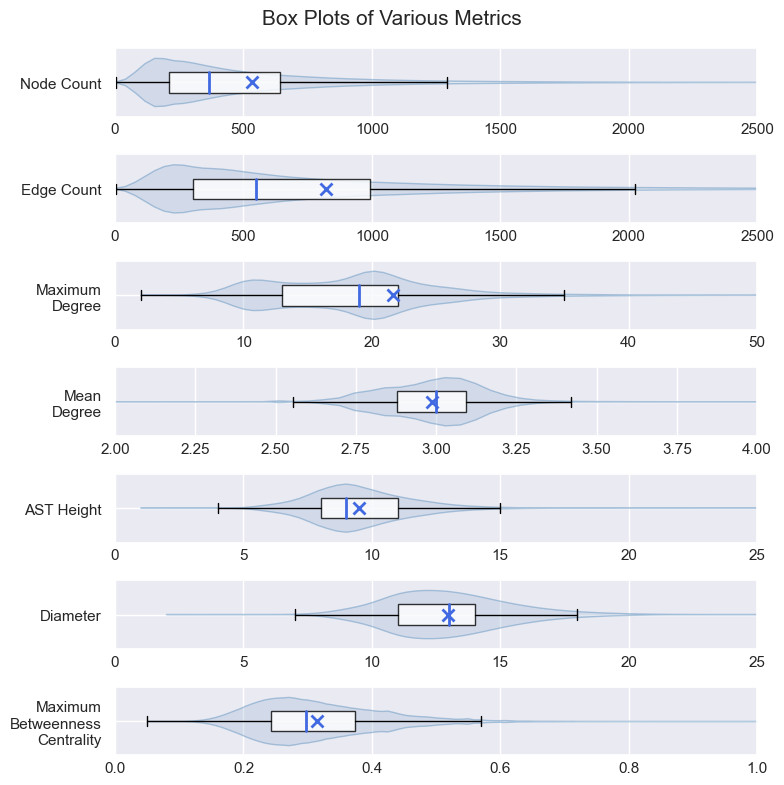}
    \caption{Box plots for various metrics of program graphs for Python submissions in Project CodeNet. The vertical blue line in each boxplot shows the median of the data as usual, while the blue $\times$ shows the mean.}
    \label{fig:boxplots}
\end{minipage}
\end{figure}

\pg cannot construct graph representations for every submission in Project CodeNet, as many of them do not parse.
Table~\ref{tab:graph-status} shows how many of the programs graph construction succeeds for, as well as the failure reasons for the remaining graphs.
The programs for which \pg cannot produce graph representations are predominantly those which fail to parse under Python's own parser: \mintinline{python}{ast.parse}. The majority of such programs cause the parser to raise a SyntaxError, IndentationError, or TabError, with just nine leading the built in parser to raise a RecursionError or ValueError.
The \pg library rejects an additional 2100 programs (0.07\%) because they contain either \mintinline{python}{return} outside of a function body, or \mintinline{python}{break} or \mintinline{python}{continue} outside of a loop. In total, this result gives us confidence there are no language feature corner cases that elude the \pg library and cause failures for well-formed programs that otherwise can be run by a standard Python interpreter. In Table~\ref{tab:type-frequencies} we report for each program graph edge type the fraction of programs it appears in as well as the mean number of appearances across all programs.

We next use \pg to measure the cyclomatic complexity of each of the submissions. Figure~\ref{fig:cyclomatic-length} plots the relationship between program length and cyclomatic complexity. We measure program length in non-empty lines of code (LOC). Omitting as outliers those programs longer than 800 LOC or with complexity exceeding 200 (just 118 programs out of 3.16 million), we perform linear regression and observe $R^2 = 0.896$, in line with prior work \citep{empirical-java-metrics, complexity-metrics}.

We measure the size of program graphs according to their node counts and edge counts, the height of their AST backbone, and graph diameter. As measures of connectedness, we compute the maximum degree of a node, mean degree of the nodes, and maximum betweenness centrality of a node in the graph.
The distributions of each of these metrics are shown with boxplots in Figure~\ref{fig:boxplots}, and key summary statistics are listed in Table~\ref{tab:metric-statistics}. Appendix~\ref{app:metrics-histograms} contains histograms showing the distribution of each metric across the dataset.
These metrics convey the scale and diversity of submissions to online programming contests and the graph sizes needed for processing these submissions as python graphs with graph neural networks.

\begin{table}[t!]
    \centering
    \begin{tabular}{lrrrr}
        \toprule
        \textbf{Metric}                  & \multicolumn{1}{c}{\textbf{Min}} & \multicolumn{1}{c}{\textbf{Median}} & \multicolumn{1}{c}{\textbf{Mean}} & \multicolumn{1}{c}{\textbf{Max}} \\
    \midrule
        Node Count                       & 3\phantom{.0}    & 364\phantom{.0}  & 534.8            & 751,817\phantom{.0} \\
        Edge Count                       & 2\phantom{.0}    & 548\phantom{.0}  & 822.4            & 4,675,600\phantom{.0} \\
        Maximum Degree                   & 2\phantom{.0}    & 19\phantom{.0}   & 21.6             & 150,004\phantom{.0} \\
        Mean Degree                      & 1.3              & 3.0              & 3.0              & 79.9             \\
        AST Height                       & 1\phantom{.0}    & 9\phantom{.0}    & 9.5              & 269\phantom{.0}  \\
        Diameter                         & 2\phantom{.0}    & 13\phantom{.0}   & 13.0             & 143\phantom{.0}  \\
        Maximum Betweenness Centrality   & 0.0              & 0.3              & 0.3              & 1.0              \\
        \bottomrule
    \end{tabular}
    \caption{Summary statistics for various graph metrics across the dataset. The diameter and maximum betweenness centrality metrics do not include graphs exceeding 5000 nodes.}
    \label{tab:metric-statistics}
\end{table}

\section{Discussion}

The core capabilities of \pg for machine learning research are generating control-flow graphs, performing data-flow analyses, generating composite program graphs, and computing cyclomatic complexity of Python programs.
For our research, we have been fruitfully using \pg for graph representations of programs for multiple years.
The library is robust and flexible, having been successfully run on millions of programs and used in several published papers.
Still, several open challenges remain for providing insights into program semantics to machine learners.
First, due to the dynamic nature of Python the library's analyses are limited to providing best-effort results, not considering the possible effects of e.g. dynamic execution or introspection.
A further key limitation of the library is its restriction to processing Python programs.
This makes getting a consistent graph representation across programming languages challenging, which is important when training a multi-lingual model of code.
While significant recent progress has been made in machine learning for code research, many fundamental problems in the space remain open research challenges.
Examples of these challenges include learning about program semantics from end-to-end program behavior, and identifying neural models exhibiting systematic generalization.
For these challenges, where the structure and semantics of programs are important, \pg provides a framework to study how graph representations of programs may contribute to forward progress.

\clearpage
\bibliographystyle{plainnat}
\bibliography{references}

\clearpage
\appendix

\section{Program Graph Visualizations}
\label{app:program-graph-examples}

\begin{figure}[h]
\centering
\includegraphics[width=\textwidth]{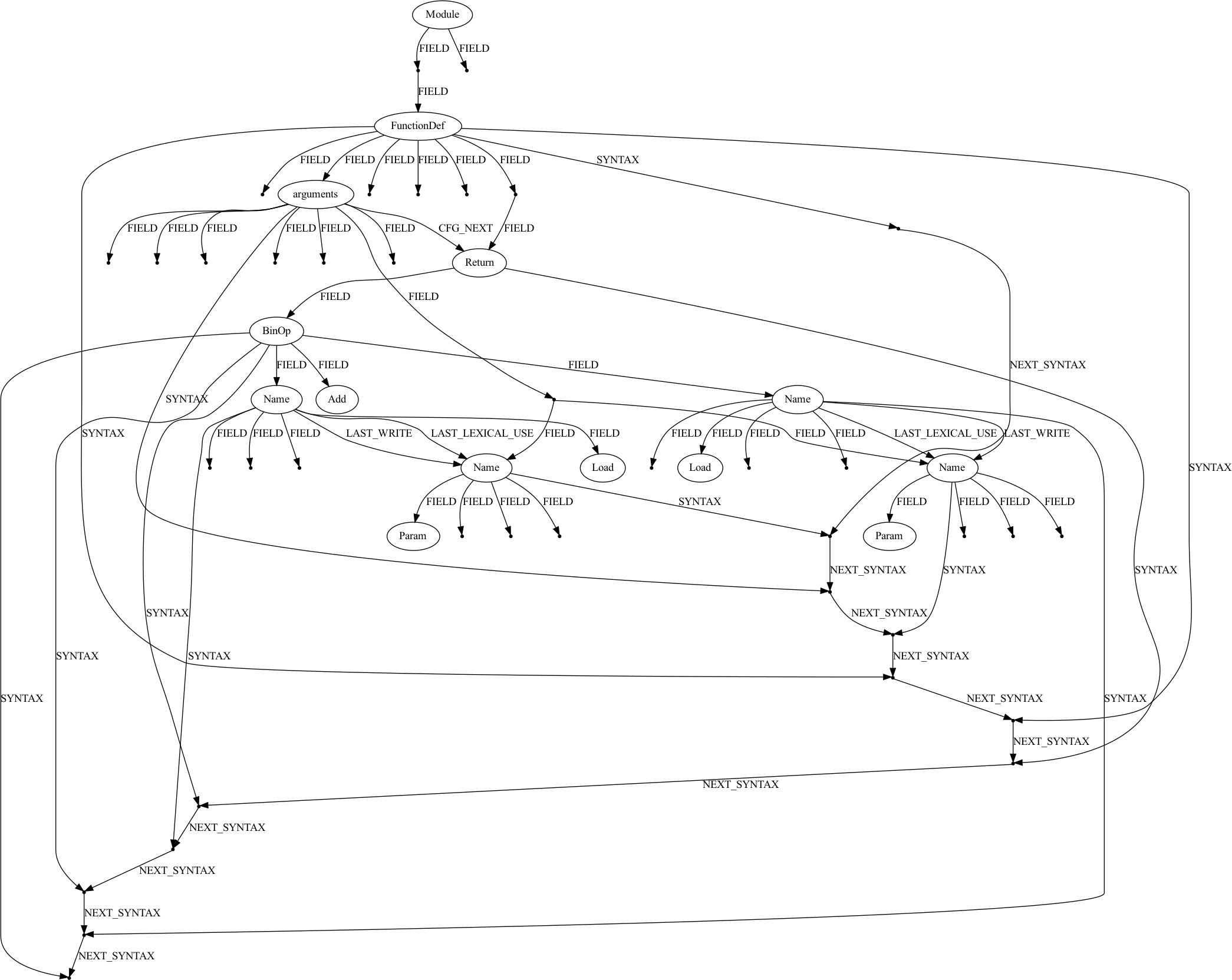}
  \caption{Program graph for Program~\#1 from Table~\ref{tab:table-of-graphs}.}
  \label{fig:program-graph-1}
\end{figure}

\begin{figure}[h]
\centering
\includegraphics[width=\textwidth]{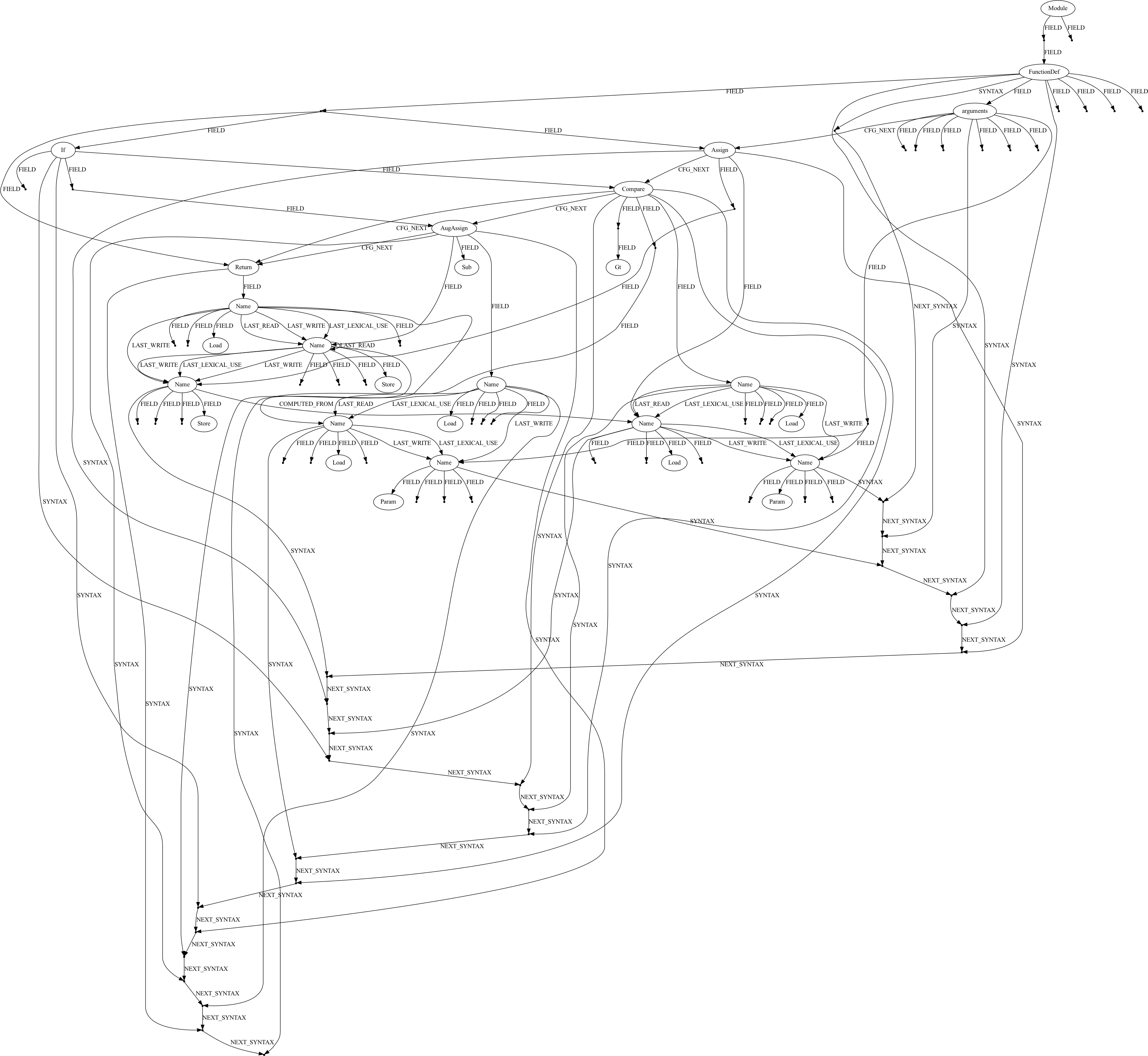}
  \caption{Program graph for Program~\#2 from Table~\ref{tab:table-of-graphs}.}
  \label{fig:program-graph-2}
\end{figure}

\begin{figure}[h]
\centering
\includegraphics[width=\textwidth]{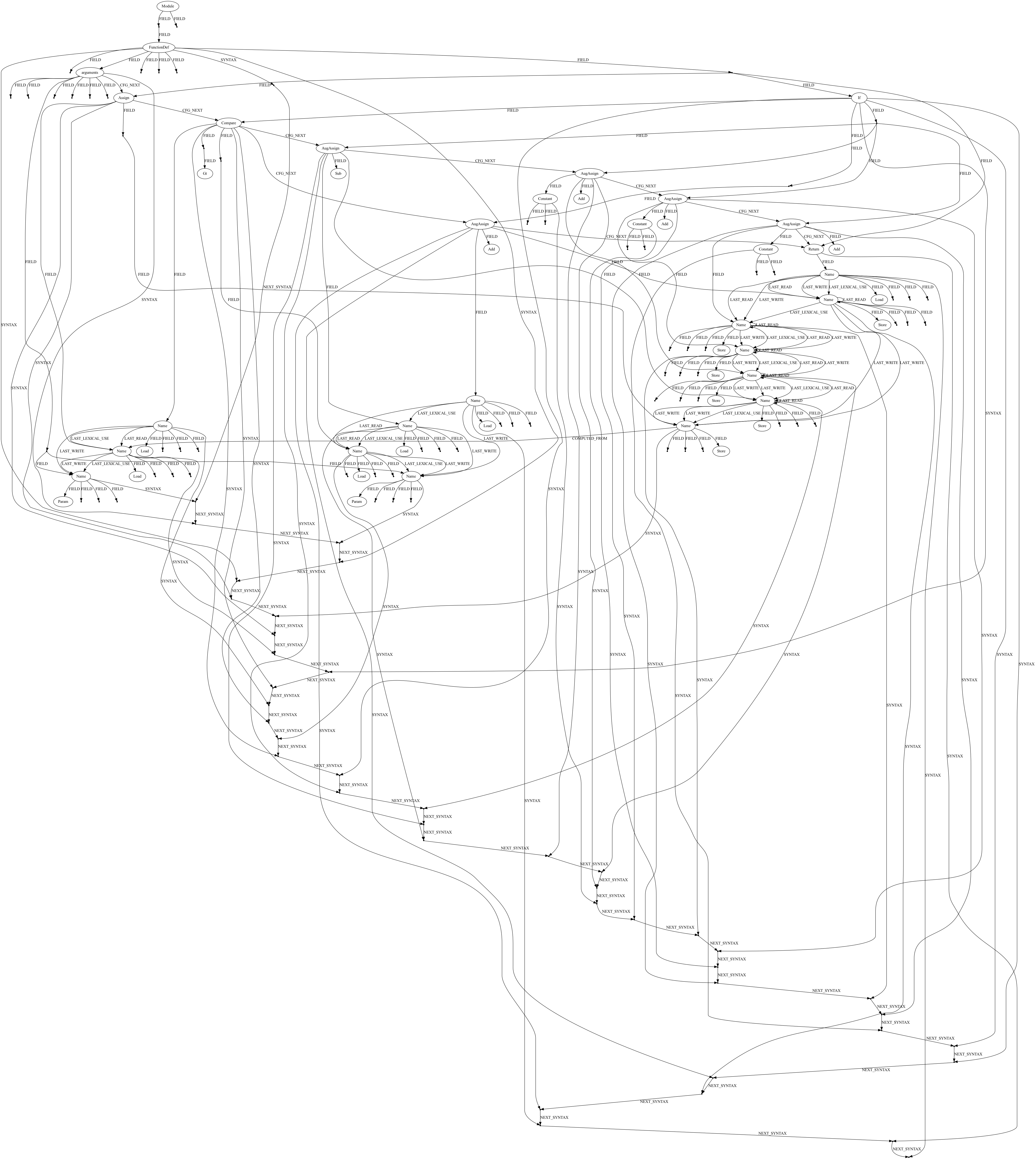}
  \caption{Program graph for Program~\#3 from Table~\ref{tab:table-of-graphs}.}
  \label{fig:program-graph-3}
\end{figure}

\begin{figure}[h]
\centering
\includegraphics[width=\textwidth]{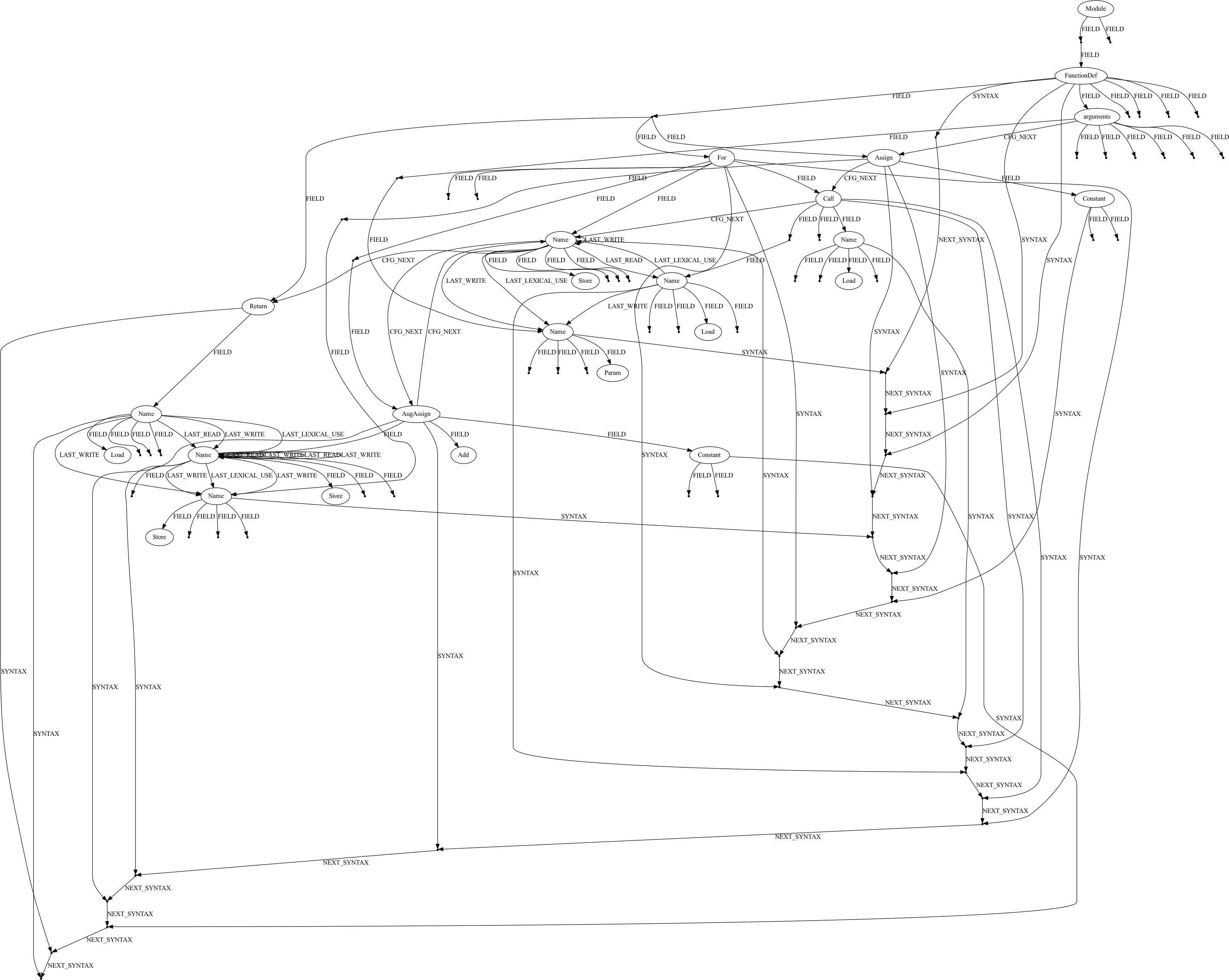}
  \caption{Program graph for Program~\#4 from Table~\ref{tab:table-of-graphs}.}
  \label{fig:program-graph-4}
\end{figure}

\begin{figure}[h]
\centering
\includegraphics[width=\textwidth]{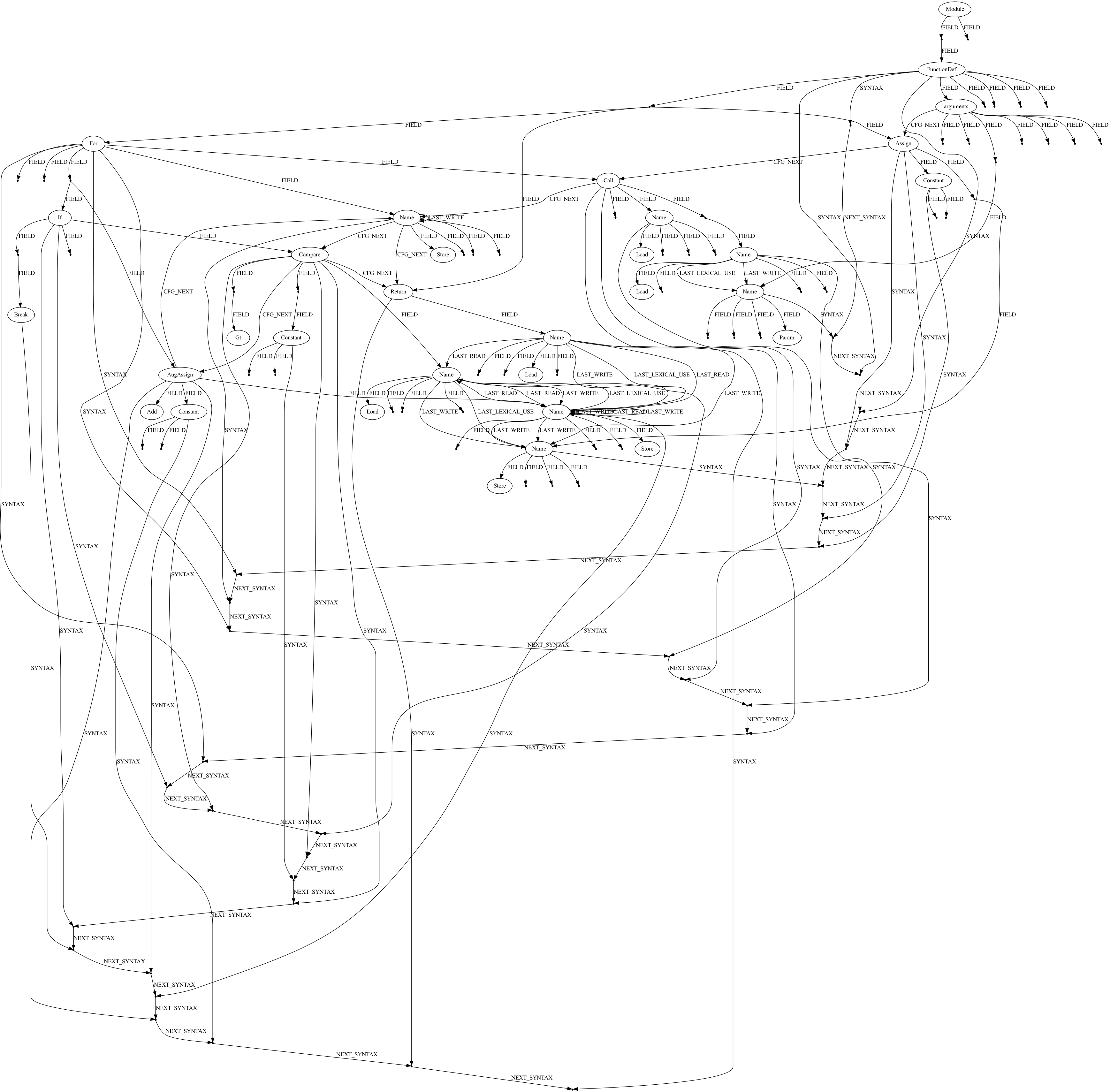}
  \caption{Program graph for Program~\#5 from Table~\ref{tab:table-of-graphs}.}
  \label{fig:program-graph-5}
\end{figure}

\begin{figure}[h]
\centering
\includegraphics[width=\textwidth]{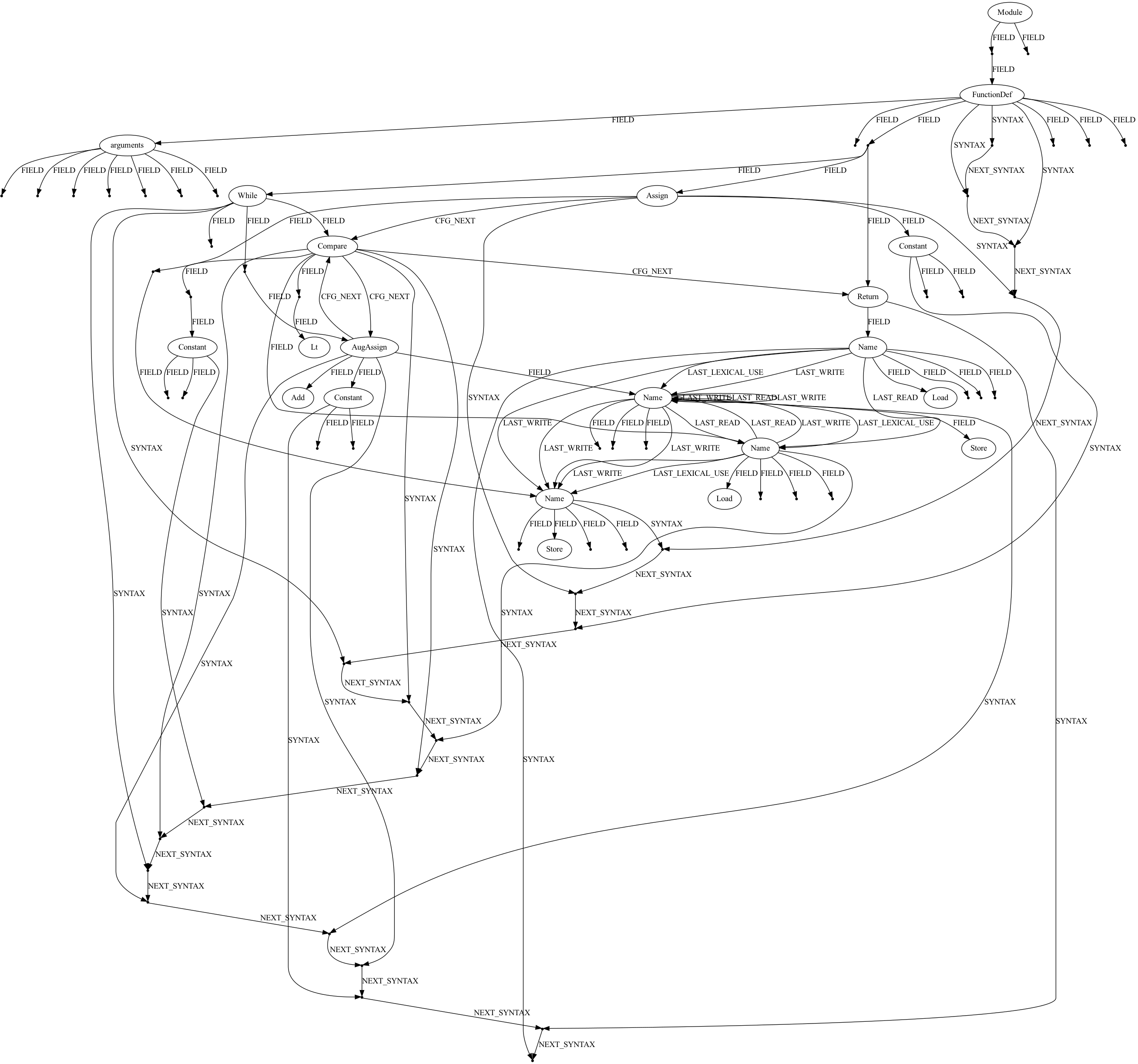}
  \caption{Program graph for Program~\#6 from Table~\ref{tab:table-of-graphs}.}
  \label{fig:program-graph-6}
\end{figure}

\begin{figure}[h]
\centering
\includegraphics[width=\textwidth]{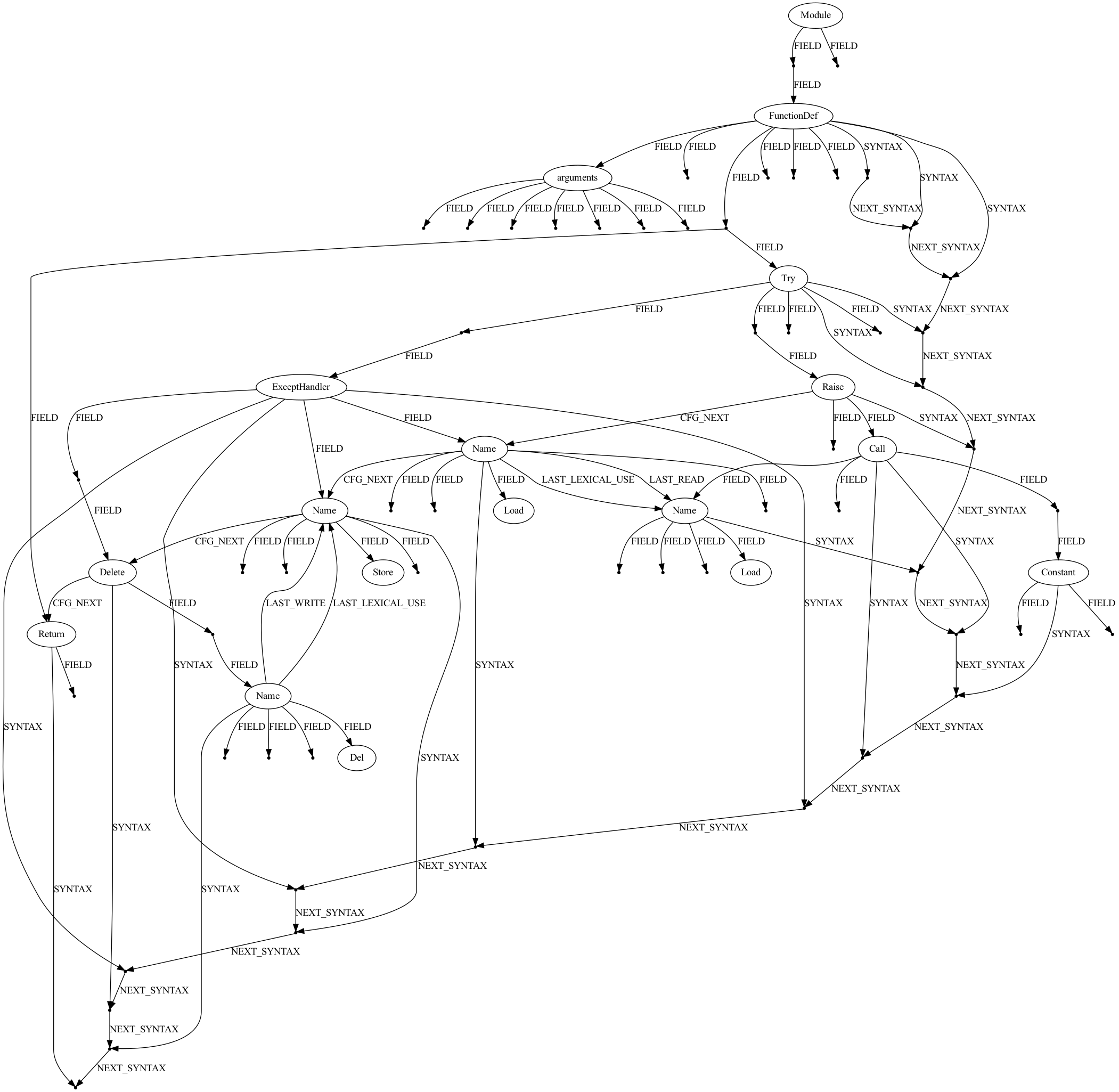}
  \caption{Program graph for Program~\#7 from Table~\ref{tab:table-of-graphs}.}
  \label{fig:program-graph-7}
\end{figure}

\begin{figure}[h]
\centering
\includegraphics[width=\textwidth]{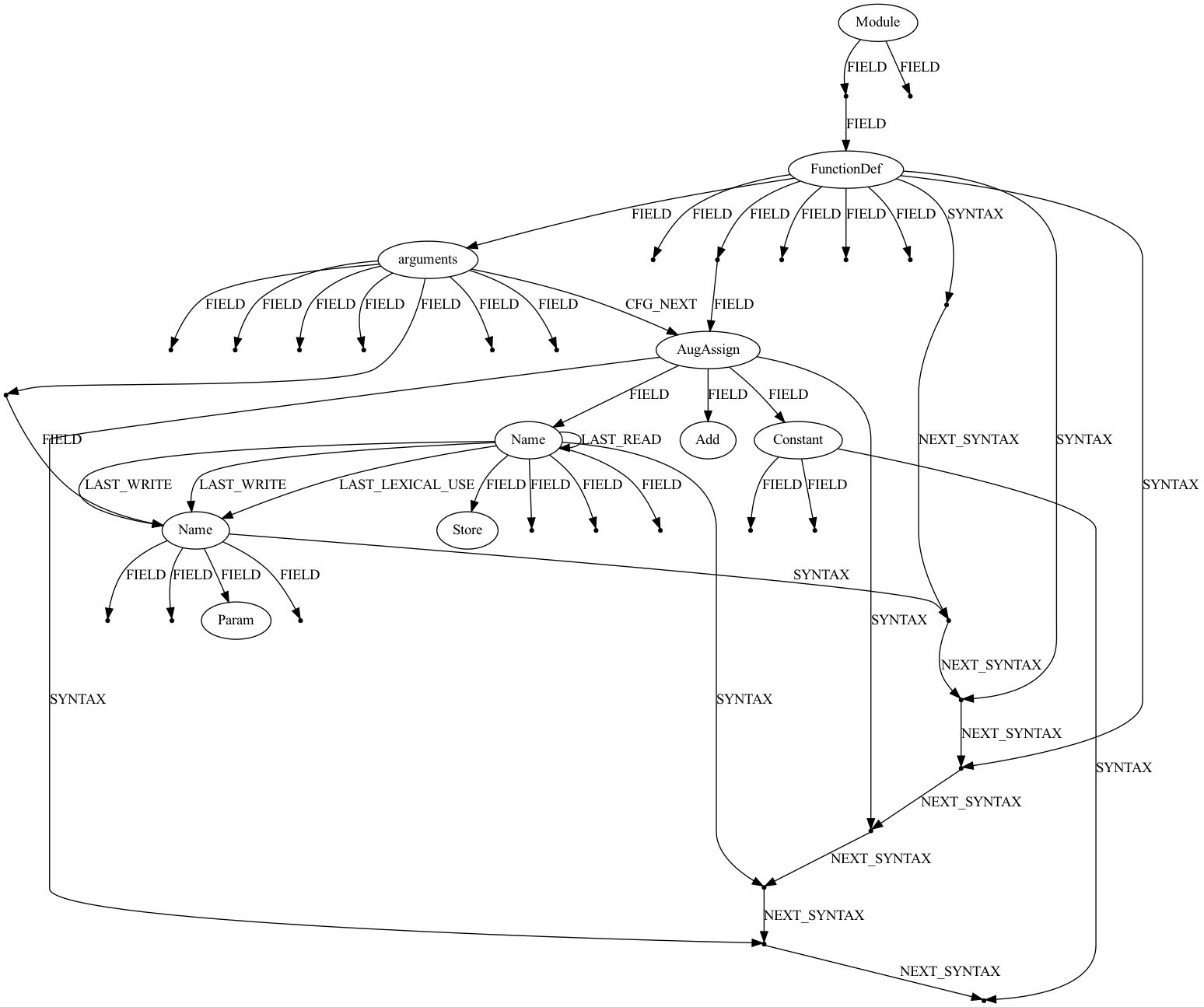}
  \caption{Program graph for Program~\#8 from Table~\ref{tab:table-of-graphs}.}
  \label{fig:program-graph-8}
\end{figure}

\clearpage
\section{Histograms of Program Graph Metrics}
\label{app:metrics-histograms}

\begin{figure}[h!]
\centering
\subfloat{
    \includegraphics[width=0.48\textwidth]{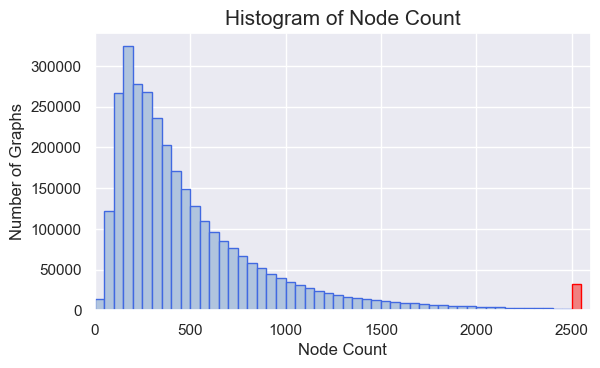}
}%
\hspace{0.01\textwidth}%
\subfloat{
  \includegraphics[width=0.48\textwidth]{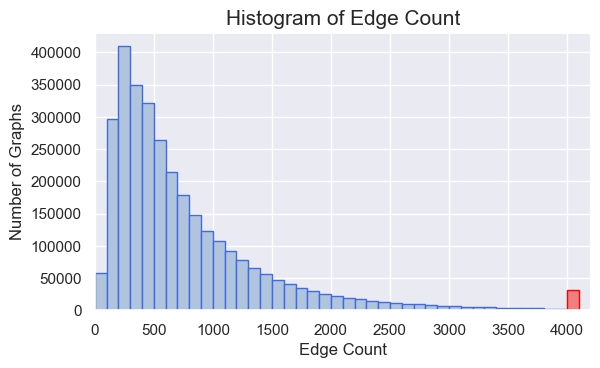}
}
\vspace{0.5em}
\subfloat{
  \includegraphics[width=0.48\textwidth]{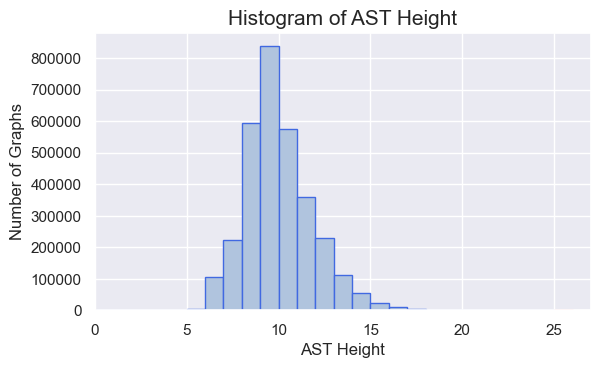}
}%
\hspace{0.01\textwidth}%
\subfloat{
  \includegraphics[width=0.48\textwidth]{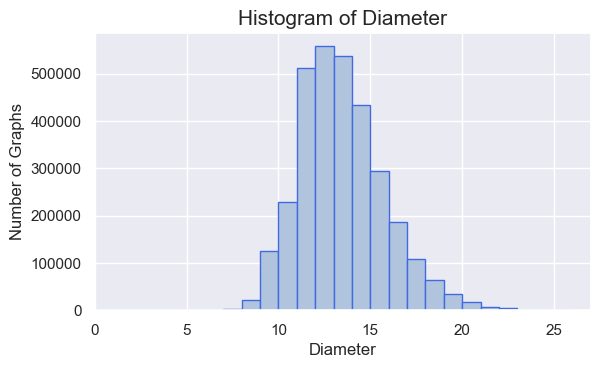}
}
\vspace{0.5em}
\subfloat{
  \includegraphics[width=0.48\textwidth]{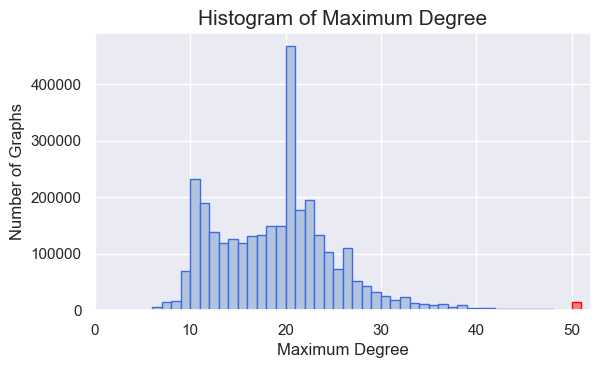}
}%
\hspace{0.01\textwidth}%
\subfloat{
  \includegraphics[width=0.48\textwidth]{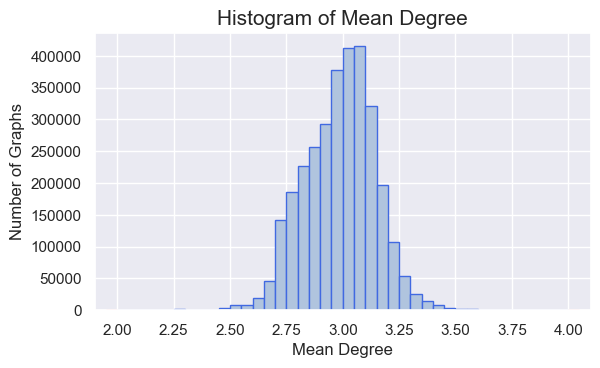}
}
\vspace{0.5em}
\subfloat{
  \includegraphics[width=0.48\textwidth]{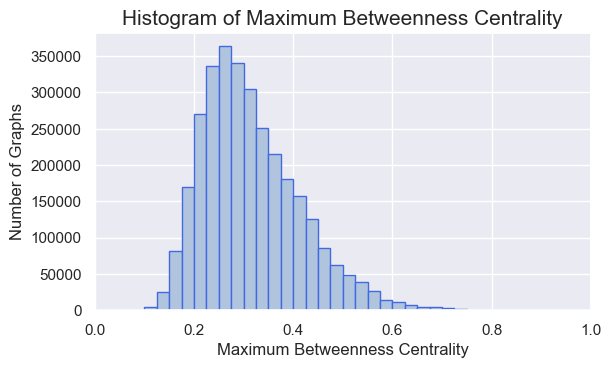}
}
\vspace{1em}
\caption{Histograms for various metrics of program graphs across the Project CodeNet dataset. Red bars include the program graphs for which the metric falls outside the range covered by the other bars.}
\end{figure}

\end{document}